\documentclass[twoside]{article}

\usepackage[accepted]{aistats2021}

\usepackage[round,authoryear]{natbib}

\usepackage{xr-hyper}
\usepackage{hyperref}
\usepackage{url}
\usepackage{booktabs}
\usepackage{amsfonts}
\usepackage{nicefrac}
\usepackage{microtype}
\usepackage{amsmath,amssymb}
\usepackage{subcaption}
\usepackage{graphicx}
\usepackage{caption}
\usepackage{titlesec}
\usepackage{color}
\usepackage{colortbl}
\usepackage{mathrsfs}
\usepackage[ruled,vlined]{algorithm2e}
\usepackage{tikz}
\usepackage{amsthm}
\usepackage{graphicx}
\usepackage{threeparttable}
\usepackage{multirow}
\usepackage{xcolor}

\usetikzlibrary{patterns}
\usetikzlibrary{calc}

\definecolor{darkteal}{HTML}{1697B7}
\definecolor{darkred}{HTML}{C23000}
\definecolor{darkblue}{rgb}{0,0.08,0.45}
\hypersetup{
    colorlinks=true,
    citecolor=darkblue,
    linkcolor=darkred,
    urlcolor=darkteal
}

\newcommand{\rad}{0.15}

\newcommand{\vsvdsize}{0.6}
\newcommand{\httsize}{1.3}
\newcommand{\vttsize}{0.6}
\newcommand{\din}{\mathrm{in}}
\newcommand{\dout}{\mathrm{out}}
\newcommand{\rpm}{\raisebox{.3ex}{$\scriptstyle\pm$}}
\newcommand{\diag}{\mathop{\mathrm{diag}}}
\def\Plus{\texttt{+}}
\def\Minus{\texttt{-}}
\newtheorem{prop}{Proposition}

\begin{document}

\runningtitle{Spectral Tensor Train Parameterization of Deep Learning Layers}
\runningauthor{Obukhov, Rakhuba, Liniger, Huang, Georgoulis, Dai, Van Gool}

\twocolumn[

\aistatstitle{Spectral Tensor Train Parameterization of Deep Learning Layers}

\aistatsauthor{ 
  Anton Obukhov \\
  ETH Zurich
    \And
  Maxim Rakhuba \\
  HSE University
    \And
  Alexander Liniger \\
  ETH Zurich
    \AND
  Zhiwu Huang \\
  ETH Zurich
    \And
  Stamatios Georgoulis \\
  ETH Zurich
    \And
  Dengxin Dai \\
  ETH Zurich
    \And
  Luc Van Gool \\
  ETH Zurich, KU Leuven
}

\aistatsaddress{ 
} 

]

\begin{abstract}
  We study low-rank parameterizations of weight matrices with embedded spectral properties in the Deep Learning context.
  The low-rank property leads to parameter efficiency and permits taking computational shortcuts when computing mappings. 
  Spectral properties are often subject to constraints in optimization problems, leading to better models and stability of optimization.
  We start by looking at the compact SVD parameterization of weight matrices and identifying redundancy sources in the parameterization. 
  We further apply the Tensor Train (TT) decomposition to the compact SVD components, and propose a non-redundant differentiable parameterization of fixed TT-rank tensor manifolds, termed the Spectral Tensor Train Parameterization (STTP).
  We demonstrate the effects of neural network compression in the image classification setting and both compression and improved training stability in the generative adversarial training setting.
  Project website: \href{https://www.obukhov.ai/sttp}{obukhov.ai/sttp}.
\end{abstract}

\section{Introduction}

Deep neural networks have become ubiquitous over the past decade in many computer science domains such as computer vision~\citep{alexnet} and natural language processing~\citep{attentionisallyouneed}. 
Much of the research was dedicated to improving model performance on various datasets and benchmarks, which has led to models with billions of parameters.
On the other hand, research productization has led to advances in model compression and energy efficiency, required by constrained computational environments such as edge devices.
Adapting research models for production is a challenging task, often involving a ground-up redesign of the model architecture, as seen in~\cite{howard2017mobilenets}.
Changes to the model often lead to a vastly different optimization landscape, which may present an additional challenge, especially in unstable settings, such as Generative Adversarial Networks (GAN)~\citep{goodfellow2014generative}.
Therefore, there is a demand for parameter-efficient drop-in neural network components (such as the linear and convolutional layers) with variable capacity and improved training stability for a wide range of optimization settings.

To tackle these challenges, we introduce a principled way to construct low-rank convolutional and linear layers with embedded spectral properties through weight matrix reparameterization. 
The usage of low-rank layers in place of the original ones introduces network compression in terms of the number of parameters.
A layer rank can be treated as a hyperparameter, which defines the layer capacity and its computational cost, and does not affect the layer dimensions.
Embedded spectral properties permit efficient rank utilization within the layer and prevent the growth of the layer's Lipschitz constant during training, which improves the optimization process stability and the final model performance.

To this end, we propose two parameterizations, which represent a weight matrix as a product of the compact SVD components: $W=U\Sigma V^\top$. 
In SVD Parameterization (SVDP), we directly parameterize $\Sigma$ using free parameters, and $U,V$ using a parameterization of orthonormal frames, such as the Householder parameterization. 
Fixing or penalizing $\Sigma$ corresponds to embedding spectral properties into the layer.
$W$ is differentiable with respect to parameters of the components, and hence parameter gradients can be computed using auto-differentiation and updated using a standard optimizer such as SGD. 
Next, we propose a Spectral Tensor Train Parameterization (STTP), further representing $U$ and $V$ through several parameterizations of much smaller orthonormal frames. 
STTP introduces sparsity into $U$ and $V$, leading to fewer parameters than SVDP with the same rank. 
It is worth noting that both proposed parameterizations of weight matrices are non-redundant.
To the best of our knowledge, the case of a differentiable non-redundant parameterization of fixed TT-rank tensor manifolds is a novel result.
From the practical point of view, given the same budget of parameters, STTP spans a different submanifold of weight matrices than SVDP, which results in more expressive layers in certain rank ranges.

Our scenario of interest includes following a pre-defined training protocol without transfer learning, with a few modifications. 
Before the training begins, we replace selected layers with low-rank ones of compatible dimensions. 
During the training, an optional spectral penalty is added to the loss function before backpropagation. 
After the training, the network can be stored in the parameterized form, or the layers can be decompressed into the original convolutional and linear types.

The paper is structured as follows. Sec.~\ref{sec:prelim} defines the notation and key terms.
SVDP is introduced in Sec.~\ref{sec:method}. 
The parameterization of orthonormal components arising from SVDP is discussed in Sec.~\ref{sec:param_stiefel_full}. 
Sec.~\ref{sec:identityspectrum} highlights parameter redundancy in SVDP and proposes a non-redundant modification. 
We introduce STTP in Sec.~\ref{sec:param_stiefel_tt} and reuse the results from Sec.~\ref{sec:identityspectrum} to remove the redundancy. 
Sec.~\ref{sec:spectralconstraints} discusses the spectral constraints applicable in both SVDP and STTP. 
We compare parameterizations and spectral constraints in the context of training GAN and image classification networks in Sec.~\ref{sec:exp}. 
Sec.~\ref{lbl:conclusion} concludes the paper.

\textbf{Related Work}\,\,
\cite{zhang2018stabilizing} explore SVD and Householder parameterizations in the context of vanishing gradients in the transition matrix of recurrent neural networks (RNN) and study representation power and generalization bounds of spectral RNN layers.
Although other orthogonal parameterization approaches exist, such as exponential maps~\citep{lezcano2019cheap} and Givens rotations, Householder transformation is found to be the most efficient~\citep{shepard2015representation}. 
The Tensor Train (TT) decomposition by~\cite{oseledets2011tensor} is used to parameterize weight matrices in a low-rank fashion in~\cite{ttrnn,garipov2016ultimate,tensorizing}. 
Both convolutional and linear layers are shown to have an adequate low-rank parameterization, although with no regard to the spectral properties or the redundancy of the proposed parameterizations.
Other low-rank tensor parameterizations have been used for network compression~\citep{obukhov2020tbasis, wang2018wide,speedingcp}, offering high compression rates at the cost of undefined spectral properties and representation redundancy.
\cite{kanakis2020reparameterizing} propose an SVD-based initialization for multitask learning.
\cite{10.1007/978-3-030-58526-6_31} address instabilities arising during CP decomposition of weight matrices during training.
\cite{holtz2012manifolds} provide the exact dimensionality of fixed TT-rank tensor manifolds; however, parameterizations are not discussed.
Despite the similar naming of \cite{sttd}, their paper is concerned with spectral approximation theory and extending TT to functions of continuous variables.
Finally, spectrum control effectively addresses multiple neural network training problems such as representation degeneration~\citep{Wang2020Improving} and mode collapse~\citep{miyato2018spectral}.

\section{Preliminaries}
\label{sec:prelim}

We are concerned with the class of neural network models $\mathcal{F}_{\theta}$ with learned parameters $\theta$ composed of affine and non-linear mappings. 
For example, a feed-forward network with $L$ layers takes the form $\omega_L \circ  a_{L-1} \circ \omega_{L-1} \circ \dots \circ a_{1} \circ \omega_{1}$, where $\omega_k$ are learned affine mappings, $a_k$ are non-linear mappings (activations), and $\circ$ denotes composition, meaning that the output of an $i$-th layer is the input of the $(i+1)$-th layer.  

The \textit{Lipschitz constant} of a mapping $\mu: \mathbb{R}^{M} \to \mathbb{R}^{N}$ is such a constant $K_{\mu}$ (if it exists) that the inequality $\|\mu(x) - \mu(y)\|_2 \leq K_{\mu} \|x - y\|_2$ holds for any $x,y \in \mathbb{R}^{M}$. 
Most non-linearities (such as ReLU, sigmoid, etc.) have their Lipschitz constant equal to 1. 
If every layer of the feed-forward model is Lipschitz-continuous, so is the composition of the layers, and thus the upper bound of the Lipschitz constant of such network is given by:
\begin{equation}\label{eq:lipsch}
    K_{\mathcal{F}_{\theta}} \leq \prod_{\ell=1}^{L} K_{\omega_{\ell}}.
\end{equation}
Similar bounds can be derived for most computational graphs corresponding to popular deep architectures such as CNNs, RNNs, Transformers, and others.

A \textit{linear layer} is an affine mapping: $\omega_{\ell}\colon \mathbb{R}^{d_{\ell}} \to \mathbb{R}^{d_{\ell+1}} : x \mapsto W_{\ell} x + b_\ell,$ with $W_{\ell}\in \mathbb{R}^{d_{\ell+1}\times d_\ell}$ (weight matrix), $b_\ell\in \mathbb{R}^{d_{\ell+1}}$ (bias term), $\ell=1,\dots,L$. 
The Lipschitz constant of a linear layer is equal to the largest singular value of the layer's weight matrix: $K_{\omega_{\ell}}=\sigma_1(W_{\ell})$.

An $N$-dimensional \textit{convolutional layer} acting in spatial dimensions $S_1, \dots, S_N$ is an affine mapping: $\omega\colon \mathbb{R}^{C_{\mathrm{in}} S_1\dots S_N} \to \mathbb{R}^{C_{\mathrm{out}} S_1\dots S_N}$, given by the kernel tensor $\mathcal{W} \in \mathbb{R}^{C_{\mathrm{out}}\times C_{\mathrm{in}} \times M_1 \times \dots \times M_N}$. Following\footnote{
  The Lipschitz constant of a convolutional layer may be larger than the largest singular value of the kernel matrix, as noted in~\cite{SedghiGL19}. 
  Nevertheless, the empirical observations in~\cite{sanyal2019stable} suggest that it does not often happen in practice.
} the conventions set by~\cite{miyato2018spectral} when dealing with convolutional layers, we are concerned with the kernel tensor reshaped into a kernel matrix of size $C_{\mathrm{out}} \times (C_{\mathrm{in}} \cdot M_1 \cdots M_N)$, also called the weight matrix.

\textit{TT decomposition}~\citep{oseledets2011tensor} is a representation for a low-rank approximation of an arbitrary $D$-dimensional array (tensor) $A \in \mathbb{R}^{n_1 \times \dots \times n_D}$ through several three-dimensional tensors (TT-cores) $\mathcal{C}^{(i)} \in \mathbb{R}^{R_{i-1} \times n_i \times R_i}, i=1,\dots,D$, with TT-rank $(R_0,\dots,R_D)$. 
Its elements are expressed as follows:
\begin{equation}
\label{eq:ttprelim}
    A_{i_1, \dots, i_D} = \!\!\!\!\!\! \sum_{\beta_0,\dots,\beta_D=1}^{R_0,..,R_D} \!\!\!
    \mathcal{C}_{\beta_0, i_1,\beta_1}^{(1)} \!\! \cdot
    \mathcal{C}_{\beta_1,i_2,\beta_2}^{(2)} \!\! \cdot\!\cdot\!\cdot \,
    \mathcal{C}_{\beta_{D\text{-}1}, i_{D}, \beta_{D}}^{(D)}
\end{equation}
The \textit{TT-rank} $R$ defines the degree of compression of $A$.
By convention, $R_0=R_D=1$, and the rest of the rank values are bounded~\citep[Eq.~(20)]{holtz2012manifolds}:
\begin{equation}
\label{eq:ttrankcapprelim}
    1 \leq R_k \leq R_k^{\mathrm{max}} \equiv  \min \left( \prod_{j=1}^k n_j, \,\negthickspace \prod_{j=k+1}^D n_j \right).
\end{equation}
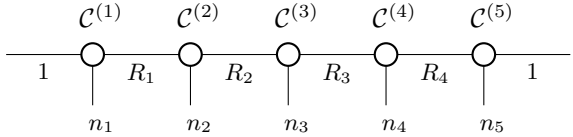
\begin{figure}[t]
\centering
\begin{tikzpicture}
\foreach \i in {1,...,7}
{
    \node (p\i) at (\i*\httsize,0) {};
}
\path[-] ($(p1) + (\rad,0)$) edge node[anchor=center, below] {{\footnotesize $1$}} ($(p2) - (\rad,0)$);
\foreach \i in {2,...,6}
{
    \pgfmathtruncatemacro{\ii}{\i + 1}
    \pgfmathtruncatemacro{\iim}{\i - 1}
    \ifthenelse{\i < 6}{
        \path[-] ($(p\i) + (\rad,0)$) edge node[anchor=center, below] {{\footnotesize $R_{\iim}$}} ($(p\ii) - (\rad,0)$);
    }
    {
        \path[-] ($(p\i) + (\rad,0)$) edge node[anchor=center, below] {{\footnotesize $1$}} ($(p\ii) - (\rad,0)$);
    }
    \draw[thick] (p\i) circle (\rad);
    \path[-] ($(p\i) - (0,\rad)$) edge ($(p\i) - (0,0.5*\rad+\vttsize)$);
    \node at ($(p\i) - (0,\rad+\vttsize+0.2)$) {\ \ {\small $n_{\iim}$}};
    \node at ($(p\i) + (0,3.5*\rad)$) {\ \ $\mathcal{C}^{(\iim)}$};
}
\end{tikzpicture}
\caption[]{
A tensor diagram corresponding to the TT decomposition of a 5D tensor in $\mathbb{R}^{n_1\times n_2 \times n_3 \times n_4 \times n_5}$.
}
\label{fig:ttnetwork:simple}
\end{figure}%
\begin{figure*}[ht]
\centering
\subfloat[]{%
\label{fig:svd_network}{
\begin{tikzpicture}

\node (p4) at (-\httsize,0) {};
\node (p5) at (0,0) {};
\node (p6) at (\httsize,0) {};

\node (phantom) at (0,-1.9*\vsvdsize) {};

\path[-] ($(p4) + (\rad,0)$) edge node[anchor=center, below] {{\footnotesize $r$}} ($(p5) - (\rad,0)$);
\path[-] ($(p5) + (\rad,0)$) edge node[anchor=center, below] {{\footnotesize $r$}} ($(p6) - (\rad,0)$);

\draw[fill] ($(p4) + (-45:\rad)$) arc (-45:135:\rad) -- cycle ;
\draw[fill] (p5) -- ($(p5) + (0:\rad)$) arc (0:360:\rad) -- cycle ;
\draw[fill] ($(p6) + (45:\rad)$) arc (45:225:\rad) -- cycle ;

\draw[thick] (p4) circle (\rad);
\draw[thick] (p5) circle (\rad);
\draw[thick] (p6) circle (\rad);

\path[-] ($(p4) - (0,\rad)$) edge ($(p4) - (0,0.5*\rad+\vttsize)$);
\path[-] ($(p6) - (0,\rad)$) edge ($(p6) - (0,0.5*\rad+\vttsize)$);

\node at ($(p4) - (0,\rad+\vttsize+0.2)$) {\ \ {\small $d_{\dout}$}};
\node at ($(p6) - (0,\rad+\vttsize+0.2)$) {\ \ {\small $d_{\din}$}};

\node at ($(p4) + (0,3.5*\rad)$) {\ $U$};
\node at ($(p5) + (0,3.5*\rad)$) {$\Sigma$};
\node at ($(p6) + (0,3.5*\rad)$) {\ $V^{\top}$};


\end{tikzpicture}
}}\qquad
\subfloat[]{%
\label{fig:tt_network}{
\begin{tikzpicture}

\foreach \i in {1,...,9}
{
    \node (p\i) at (\i*\httsize,0) {};
}

\path[-] ($(p1) + (\rad,0)$) edge node[anchor=center, below] {{\footnotesize $R_1^{\dout}$}} ($(p2) - (\rad,0)$);
\path[-] ($(p2) + (\rad,0)$) edge node[anchor=center, below] {{\footnotesize $R_2^{\dout}$}} ($(p3) - (2*\rad,0)$);
\path[-] ($(p3) + (2*\rad,0)$) edge node[anchor=center, below] {{\footnotesize $R_{D_\dout{\texttt{-}}1}^{\dout}$}} ($(p4) - (\rad,0)$);
\path[-] ($(p4) + (\rad,0)$) edge node[anchor=center, below] {{\footnotesize $r$}} ($(p5) - (\rad,0)$);
\path[-] ($(p5) + (\rad,0)$) edge node[anchor=center, below] {{\footnotesize $r$}} ($(p6) - (\rad,0)$);
\path[-] ($(p6) + (\rad,0)$) edge node[anchor=center, below] {{\footnotesize $R_{D_\din{\texttt{-}}1}^{\din}$}} ($(p7) - (2*\rad,0)$);
\path[-] ($(p7) + (2*\rad,0)$) edge node[anchor=center, below] {{\footnotesize $R_2^{\din}$}} ($(p8) - (\rad,0)$);
\path[-] ($(p8) + (\rad,0)$) edge node[anchor=center, below] {{\footnotesize $R_1^{\din}$}} ($(p9) - (\rad,0)$);

\draw[fill] ($(p1) + (-45:\rad)$) arc (-45:135:\rad) -- cycle ;
\draw[fill] ($(p2) + (-45:\rad)$) arc (-45:135:\rad) -- cycle ;
\draw[fill] ($(p4) + (-45:\rad)$) arc (-45:135:\rad) -- cycle ;
\draw[fill] (p5)-- ($(p5) + (0:\rad)$) arc (0:360:\rad) -- cycle ;
\draw[fill] ($(p6) + (45:\rad)$) arc (45:225:\rad) -- cycle ;
\draw[fill] ($(p8) + (45:\rad)$) arc (45:225:\rad) -- cycle ;
\draw[fill] ($(p9) + (45:\rad)$) arc (45:225:\rad) -- cycle ;

\draw[thick] (p1) circle (\rad);
\draw[thick] (p2) circle (\rad);
\node at (p3) {\,\smaller $\cdots$};
\draw[thick] (p4) circle (\rad);
\draw[thick] (p5) circle (\rad);
\draw[thick] (p6) circle (\rad);
\node at (p7) {\,\smaller $\cdots$};
\draw[thick] (p8) circle (\rad);
\draw[thick] (p9) circle (\rad);

\path[-] ($(p1) - (0,\rad)$) edge ($(p1) - (0,0.5*\rad+\vttsize)$);
\path[-] ($(p2) - (0,\rad)$) edge ($(p2) - (0,0.5*\rad+\vttsize)$);
\path[-] ($(p4) - (0,\rad)$) edge ($(p4) - (0,0.5*\rad+\vttsize)$);
\path[-] ($(p6) - (0,\rad)$) edge ($(p6) - (0,0.5*\rad+\vttsize)$);
\path[-] ($(p8) - (0,\rad)$) edge ($(p8) - (0,0.5*\rad+\vttsize)$);
\path[-] ($(p9) - (0,\rad)$) edge ($(p9) - (0,0.5*\rad+\vttsize)$);

\node at ($(p1) - (0,\rad+\vttsize+0.2)$) {\ \ {\small $n^{\dout}_{1}$}};
\node at ($(p2) - (0,\rad+\vttsize+0.2)$) {\ \ {\small $n^{\dout}_{2}$}};
\node at ($(p4) - (0,\rad+\vttsize+0.2)$) {\ \ {\small $n^{\dout}_{D_\dout}$}};
\node at ($(p6) - (0,\rad+\vttsize+0.2)$) {\ \ {\small $n^{\din}_{D_\din}$}};
\node at ($(p8) - (0,\rad+\vttsize+0.2)$) {\ \ {\small $n^{\din}_{2}$}};
\node at ($(p9) - (0,\rad+\vttsize+0.2)$) {\ \ {\small $n^{\din}_{1}$}};

\node at ($(p1) + (0,3.5*\rad)$) {\ \ $\mathcal{U}^{(1)}$};
\node at ($(p2) + (0,3.5*\rad)$) {\ \ $\mathcal{U}^{(2)}$};
\node at ($(p4) + (0,3.5*\rad)$) {\ \ $\mathcal{U}^{(D_\dout)}$};
\node at ($(p5) + (0,3.5*\rad)$) {$\Sigma$};
\node at ($(p6) + (0,3.5*\rad)$) {\ \ $\mathcal{V}^{(D_\din) \top}$};
\node at ($(p8) + (0,3.5*\rad)$) {\ \ $\mathcal{V}^{(2) \top}$};
\node at ($(p9) + (0,3.5*\rad)$) {\ \ $\mathcal{V}^{(1) \top}$};

\end{tikzpicture}
}}
\caption[]{
    Tensor diagrams of (a) SVDP and (b) STTP of a weight matrix $W\in\mathbb{R}^{d_{\dout}\times d_{\din}}$, where $d_{\dout}=\prod n_i^{\dout}$, $d_{\din}=\prod n_j^{\din}$.
    Half-filled nodes represent TT-cores whose matricizations ($\mathcal{M}$) are orthonormal frames (with shaded area enumerating columns). 
    Filled nodes represent diagonal matrices.
    In both SVDP and STTP, we can choose any $r$ subject to rank constraints while still controlling the spectral properties due to the exposed matrix $\Sigma$.
    STTP provides an extra degree of compression given the same rank $r$ due to the induced sparsity in $U$ and $V$.
}
\label{fig:networks}
\end{figure*}
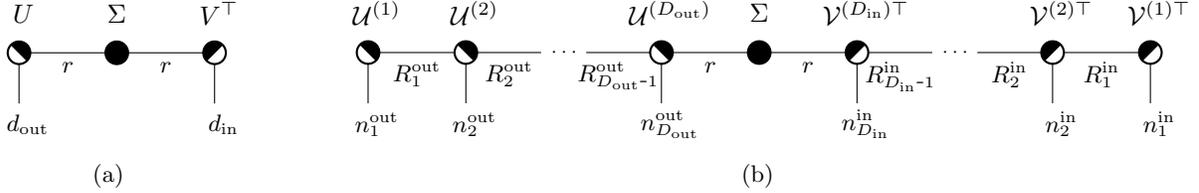%
We define the \textit{order of elements} in the tensor $A$ by associating each element $A_{i_1,\ldots,i_D}$ with a multi-index:
\begin{equation}
\label{eq:order}
\overline{i_1\dots i_D} = 1 + \sum_{p=1}^D (i_p - 1) \prod_{q=p+1}^D n_q.
\end{equation}
\textit{Reshaping} preserves the order of elements. In what follows, \textit{tensorization} refers to the reshaping of a vector or a matrix into a tensor. \textit{Matricization} (e.g., of TT-cores) refers to the reshaping of a tensor into a matrix. 

\textit{Tensor diagram notation} (Fig.~\ref{fig:ttnetwork:simple}) is a convenient tool for visualizing interactions of tensors like in~\eqref{eq:ttprelim}. 
Each node represents a tensor with the number of legs matching the number of dimensions: 1 -- vector, 2 -- matrix, 3 -- 3D array (e.g., TT-core). 
Connected legs represent summation over the corresponding indices in~\eqref{eq:ttprelim}. 
Size-1 legs may be omitted. 
\textit{Contraction} of the tensor diagram is the operation of computing the elements of the tensor product of all nodes involved in the operation (e.g., contraction of Fig.~\ref{fig:ttnetwork:simple} gives a tensor of size $n_1 \times \ldots \times n_5$).

\section{Method}
\label{sec:method}

In this section, we describe the proposed parameterization of neural network layers. Given a weight matrix $W \in \mathbb{R}^{d_\dout \times d_\din}$ of a layer implementing an affine mapping and the rank hyperparameter $r \leq \min(d_{\dout},d_{\din})$, we represent $W$ using the compact SVD with rank $r$:
\begin{equation}\label{eq:svd}
    W = U \Sigma V^\top ,
\end{equation}
where $U\in\mathbb{R}^{d_{\dout}\times r}$ and $V\in\mathbb{R}^{d_{\din}\times r}$ have orthonormal columns, and $\Sigma = \mathrm{diag}(\sigma_1,\dots, \sigma_{r})$ is a matrix of singular values, parameterized by $r$ parameters.
Matrices $U$ and $V$ belong to the real Stiefel manifold
\[
    \mathrm{St}(d, r) \equiv \{X\in\mathbb{R}^{d\times r}: X^\top X = I_r\}
\]
of orthonormal real $r$-frames ($r \leq d$), referred to as \textit{orthonormal frames}. 
In what follows, we will use $U$ of the size $d\times r$, implying either $U$ of the size $d_{\dout}\times r$ or $V$ of the size $d_{\din} \times r$, unless stated otherwise.

We directly parameterize the arising $U\in \mathrm{St}(d, r)$ by certain mappings $\phi: \mathbb{R}^q \to \mathrm{St}(d, r)$ such that $U = \phi(\theta)$ and where $q$ is the dimensionality of a submanifold of $\mathrm{St}(d,r)$, chosen according to a parameterization type.

We consider two types of weight matrix parameterizations: (1) SVDP, requiring both $U$ and $V$ parameterized as orthonormal frames (Sec.~\ref{sec:param_stiefel_full}, Fig.~\ref{fig:svd_network}), and (2) STTP, a parameterization of a reshaped weight matrix with a fixed TT-rank or, equivalently, SVDP with TT-compressed $U$ and $V$ (Sec.~\ref{sec:param_stiefel_tt}, Fig.~\ref{fig:tt_network}).

\subsection{SVDP}
\label{sec:param_stiefel_full}

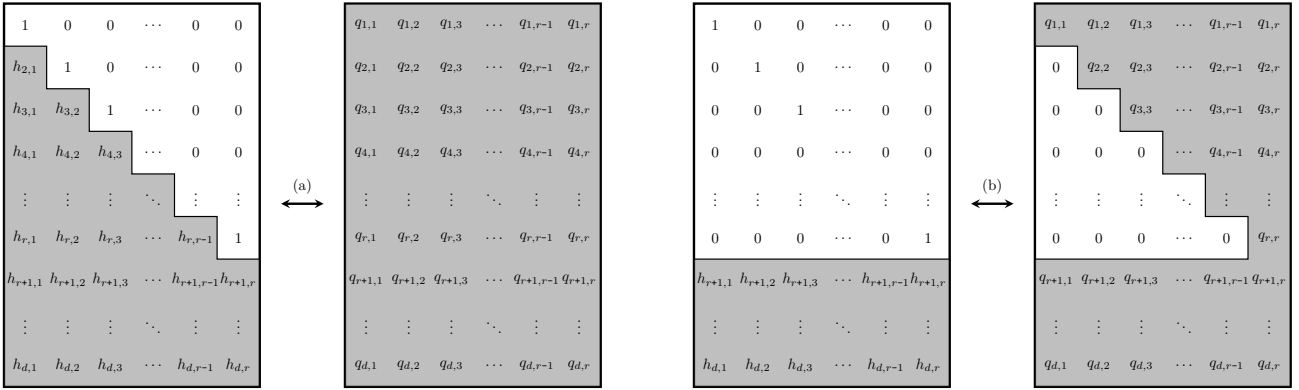
\begin{figure*}[t]
\centering
\resizebox{0.465\linewidth}{!}{%
\begin{tikzpicture}
    \draw [ultra thick, draw=black] 
        (0,9) -- (6,9) -- (6,0) -- (0,0) -- cycle;
    \draw [thick, draw=black, fill=gray, fill opacity=0.5]
       (0,8) -- (1,8) -- 
       (1,7) -- (2,7) -- 
       (2,6) -- (3,6) -- 
       (3,5) -- (4,5) -- 
       (4,4) -- (5,4) -- 
       (5,3) -- (6,3) -- 
       (6,0) -- (0,0) -- 
       cycle;

    \node[] at (0.5, 8.5) {$1$};
    \node[] at (0.5, 7.5) {$h_{2,1}$};
    \node[] at (0.5, 6.5) {$h_{3,1}$}; 
    \node[] at (0.5, 5.5) {$h_{4,1}$}; 
    \node[] at (0.5, 4.5) {$\vdots$}; 
    \node[] at (0.5, 3.5) {$h_{r, 1}$}; 
    \node[] at (0.5, 2.5) {$h_{r \Plus 1, 1}$}; 
    \node[] at (0.5, 1.5) {$\vdots$}; 
    \node[] at (0.5, 0.5) {$h_{d, 1}$}; 

    \node[] at (1.5, 8.5) {$0$};
    \node[] at (1.5, 7.5) {$1$};
    \node[] at (1.5, 6.5) {$h_{3,2}$};
    \node[] at (1.5, 5.5) {$h_{4,2}$};
    \node[] at (1.5, 4.5) {$\vdots$}; 
    \node[] at (1.5, 3.5) {$h_{r, 2}$}; 
    \node[] at (1.5, 2.5) {$h_{r \Plus 1, 2}$}; 
    \node[] at (1.5, 1.5) {$\vdots$}; 
    \node[] at (1.5, 0.5) {$h_{d, 2}$}; 

    \node[] at (2.5, 8.5) {$0$};
    \node[] at (2.5, 7.5) {$0$};
    \node[] at (2.5, 6.5) {$1$};
    \node[] at (2.5, 5.5) {$h_{4,3}$}; 
    \node[] at (2.5, 4.5) {$\vdots$}; 
    \node[] at (2.5, 3.5) {$h_{r, 3}$}; 
    \node[] at (2.5, 2.5) {$h_{r \Plus 1, 3}$}; 
    \node[] at (2.5, 1.5) {$\vdots$}; 
    \node[] at (2.5, 0.5) {$h_{d, 3}$}; 

    \node[] at (3.5, 8.5) {$\hdots$};
    \node[] at (3.5, 7.5) {$\hdots$};
    \node[] at (3.5, 6.5) {$\hdots$};
    \node[] at (3.5, 5.5) {$\hdots$};
    \node[] at (3.5, 4.5) {$\ddots$}; 
    \node[] at (3.5, 3.5) {$\hdots$}; 
    \node[] at (3.5, 2.5) {$\hdots$}; 
    \node[] at (3.5, 1.5) {$\ddots$}; 
    \node[] at (3.5, 0.5) {$\hdots$}; 

    \node[] at (4.5, 8.5) {$0$};
    \node[] at (4.5, 7.5) {$0$};
    \node[] at (4.5, 6.5) {$0$};
    \node[] at (4.5, 5.5) {$0$};
    \node[] at (4.5, 4.5) {$\vdots$}; 
    \node[] at (4.5, 3.5) {$h_{r, r \Minus 1}$}; 
    \node[] at (4.5, 2.5) {$h_{r \Plus 1, r \Minus 1}$}; 
    \node[] at (4.5, 1.5) {$\vdots$}; 
    \node[] at (4.5, 0.5) {$h_{d, r \Minus 1}$}; 

    \node[] at (5.5, 8.5) {$0$};
    \node[] at (5.5, 7.5) {$0$};
    \node[] at (5.5, 6.5) {$0$};
    \node[] at (5.5, 5.5) {$0$};
    \node[] at (5.5, 4.5) {$\vdots$}; 
    \node[] at (5.5, 3.5) {$1$}; 
    \node[] at (5.5, 2.5) {$h_{r \Plus 1, r}$}; 
    \node[] at (5.5, 1.5) {$\vdots$}; 
    \node[] at (5.5, 0.5) {$h_{d, r}$}; 

    \draw [>=stealth, <->, ultra thick, draw=black] (6.5,4.3) -- (7.5,4.3);
    \node[] at (7,4.7) {(a)};

    \pgfmathtruncatemacro{\offs}{8}

    \draw [ultra thick, draw=black] 
        (\offs+0,9) -- (\offs+6,9) -- (\offs+6,0) -- (\offs+0,0) -- cycle;
    \draw [thick, draw=black, fill=gray, fill opacity=0.5]
        (\offs+0,9) -- (\offs+6,9) -- 
        (\offs+6,0) -- (\offs+0,0) -- cycle;
       cycle;

    \node[] at (\offs+0.5, 8.5) {$q_{1,1}$};
    \node[] at (\offs+0.5, 7.5) {$q_{2,1}$};
    \node[] at (\offs+0.5, 6.5) {$q_{3,1}$}; 
    \node[] at (\offs+0.5, 5.5) {$q_{4,1}$}; 
    \node[] at (\offs+0.5, 4.5) {$\vdots$}; 
    \node[] at (\offs+0.5, 3.5) {$q_{r,1}$}; 
    \node[] at (\offs+0.5, 2.5) {$q_{r \Plus 1, 1}$}; 
    \node[] at (\offs+0.5, 1.5) {$\vdots$}; 
    \node[] at (\offs+0.5, 0.5) {$q_{d, 1}$}; 

    \node[] at (\offs+1.5, 8.5) {$q_{1,2}$};
    \node[] at (\offs+1.5, 7.5) {$q_{2,2}$};
    \node[] at (\offs+1.5, 6.5) {$q_{3,2}$};
    \node[] at (\offs+1.5, 5.5) {$q_{4,2}$};
    \node[] at (\offs+1.5, 4.5) {$\vdots$}; 
    \node[] at (\offs+1.5, 3.5) {$q_{r,2}$}; 
    \node[] at (\offs+1.5, 2.5) {$q_{r \Plus 1, 2}$}; 
    \node[] at (\offs+1.5, 1.5) {$\vdots$}; 
    \node[] at (\offs+1.5, 0.5) {$q_{d, 2}$}; 

    \node[] at (\offs+2.5, 8.5) {$q_{1,3}$};
    \node[] at (\offs+2.5, 7.5) {$q_{2,3}$};
    \node[] at (\offs+2.5, 6.5) {$q_{3,3}$};
    \node[] at (\offs+2.5, 5.5) {$q_{4,3}$}; 
    \node[] at (\offs+2.5, 4.5) {$\vdots$}; 
    \node[] at (\offs+2.5, 3.5) {$q_{r,3}$};
    \node[] at (\offs+2.5, 2.5) {$q_{r \Plus 1, 3}$}; 
    \node[] at (\offs+2.5, 1.5) {$\vdots$}; 
    \node[] at (\offs+2.5, 0.5) {$q_{d, 3}$}; 

    \node[] at (\offs+3.5, 8.5) {$\hdots$};
    \node[] at (\offs+3.5, 7.5) {$\hdots$};
    \node[] at (\offs+3.5, 6.5) {$\hdots$};
    \node[] at (\offs+3.5, 5.5) {$\hdots$};
    \node[] at (\offs+3.5, 4.5) {$\ddots$}; 
    \node[] at (\offs+3.5, 3.5) {$\hdots$}; 
    \node[] at (\offs+3.5, 2.5) {$\hdots$}; 
    \node[] at (\offs+3.5, 1.5) {$\ddots$}; 
    \node[] at (\offs+3.5, 0.5) {$\hdots$}; 

    \node[] at (\offs+4.5, 8.5) {$q_{1,r \Minus 1}$};
    \node[] at (\offs+4.5, 7.5) {$q_{2,r \Minus 1}$};
    \node[] at (\offs+4.5, 6.5) {$q_{3,r \Minus 1}$};
    \node[] at (\offs+4.5, 5.5) {$q_{4,r \Minus 1}$};
    \node[] at (\offs+4.5, 4.5) {$\vdots$}; 
    \node[] at (\offs+4.5, 3.5) {$q_{r,r \Minus 1}$}; 
    \node[] at (\offs+4.5, 2.5) {$q_{r \Plus 1, r \Minus 1}$}; 
    \node[] at (\offs+4.5, 1.5) {$\vdots$}; 
    \node[] at (\offs+4.5, 0.5) {$q_{d, r \Minus 1}$}; 

    \node[] at (\offs+5.5, 8.5) {$q_{1,r}$};
    \node[] at (\offs+5.5, 7.5) {$q_{2,r}$};
    \node[] at (\offs+5.5, 6.5) {$q_{3,r}$};
    \node[] at (\offs+5.5, 5.5) {$q_{4,r}$};
    \node[] at (\offs+5.5, 4.5) {$\vdots$}; 
    \node[] at (\offs+5.5, 3.5) {$q_{r,r}$};
    \node[] at (\offs+5.5, 2.5) {$q_{r \Plus 1, r}$}; 
    \node[] at (\offs+5.5, 1.5) {$\vdots$}; 
    \node[] at (\offs+5.5, 0.5) {$q_{d, r}$}; 

\end{tikzpicture}%
%
}
\hfill
\resizebox{0.465\linewidth}{!}{%
\begin{tikzpicture}
    \draw [ultra thick, draw=black] 
        (0,9) -- (6,9) -- (6,0) -- (0,0) -- cycle;
    \draw [thick, draw=black, fill=gray, fill opacity=0.5]
       (0,3) -- (6,3) -- 
       (6,0) -- (0,0) -- 
       cycle;

    \node[] at (0.5, 8.5) {$1$};
    \node[] at (0.5, 7.5) {$0$};
    \node[] at (0.5, 6.5) {$0$}; 
    \node[] at (0.5, 5.5) {$0$}; 
    \node[] at (0.5, 4.5) {$\vdots$}; 
    \node[] at (0.5, 3.5) {$0$}; 
    \node[] at (0.5, 2.5) {$h_{r \Plus 1, 1}$}; 
    \node[] at (0.5, 1.5) {$\vdots$}; 
    \node[] at (0.5, 0.5) {$h_{d, 1}$}; 

    \node[] at (1.5, 8.5) {$0$};
    \node[] at (1.5, 7.5) {$1$};
    \node[] at (1.5, 6.5) {$0$};
    \node[] at (1.5, 5.5) {$0$};
    \node[] at (1.5, 4.5) {$\vdots$}; 
    \node[] at (1.5, 3.5) {$0$}; 
    \node[] at (1.5, 2.5) {$h_{r \Plus 1, 2}$}; 
    \node[] at (1.5, 1.5) {$\vdots$}; 
    \node[] at (1.5, 0.5) {$h_{d, 2}$}; 

    \node[] at (2.5, 8.5) {$0$};
    \node[] at (2.5, 7.5) {$0$};
    \node[] at (2.5, 6.5) {$1$};
    \node[] at (2.5, 5.5) {$0$}; 
    \node[] at (2.5, 4.5) {$\vdots$}; 
    \node[] at (2.5, 3.5) {$0$}; 
    \node[] at (2.5, 2.5) {$h_{r \Plus 1, 3}$}; 
    \node[] at (2.5, 1.5) {$\vdots$}; 
    \node[] at (2.5, 0.5) {$h_{d, 3}$}; 

    \node[] at (3.5, 8.5) {$\hdots$};
    \node[] at (3.5, 7.5) {$\hdots$};
    \node[] at (3.5, 6.5) {$\hdots$};
    \node[] at (3.5, 5.5) {$\hdots$};
    \node[] at (3.5, 4.5) {$\ddots$}; 
    \node[] at (3.5, 3.5) {$\hdots$}; 
    \node[] at (3.5, 2.5) {$\hdots$}; 
    \node[] at (3.5, 1.5) {$\ddots$}; 
    \node[] at (3.5, 0.5) {$\hdots$}; 

    \node[] at (4.5, 8.5) {$0$};
    \node[] at (4.5, 7.5) {$0$};
    \node[] at (4.5, 6.5) {$0$};
    \node[] at (4.5, 5.5) {$0$};
    \node[] at (4.5, 4.5) {$\vdots$}; 
    \node[] at (4.5, 3.5) {$0$}; 
    \node[] at (4.5, 2.5) {$h_{r \Plus 1, r \Minus 1}$}; 
    \node[] at (4.5, 1.5) {$\vdots$}; 
    \node[] at (4.5, 0.5) {$h_{d, r \Minus 1}$}; 

    \node[] at (5.5, 8.5) {$0$};
    \node[] at (5.5, 7.5) {$0$};
    \node[] at (5.5, 6.5) {$0$};
    \node[] at (5.5, 5.5) {$0$};
    \node[] at (5.5, 4.5) {$\vdots$}; 
    \node[] at (5.5, 3.5) {$1$}; 
    \node[] at (5.5, 2.5) {$h_{r \Plus 1, r}$}; 
    \node[] at (5.5, 1.5) {$\vdots$}; 
    \node[] at (5.5, 0.5) {$h_{d, r}$}; 

    \draw [>=stealth, <->, ultra thick, draw=black] (6.5,4.3) -- (7.5,4.3);
    \node[] at (7,4.7) {(b)};

    \pgfmathtruncatemacro{\offs}{8}

    \draw [ultra thick, draw=black] 
        (\offs+0,9) -- (\offs+6,9) -- (\offs+6,0) -- (\offs+0,0) -- cycle;
    \draw [thick, draw=black, fill=gray, fill opacity=0.5]
       (\offs+0,8) -- (\offs+1,8) -- 
       (\offs+1,7) -- (\offs+2,7) -- 
       (\offs+2,6) -- (\offs+3,6) -- 
       (\offs+3,5) -- (\offs+4,5) -- 
       (\offs+4,4) -- (\offs+5,4) -- 
       (\offs+5,3) -- (\offs+0,3) -- 
       (\offs+0,0) -- (\offs+6,0) -- 
       (\offs+6,9) -- (\offs+0,9) -- 
       (\offs+0,8) --
       cycle;

    \node[] at (\offs+0.5, 8.5) {$q_{1,1}$};
    \node[] at (\offs+0.5, 7.5) {$0$};
    \node[] at (\offs+0.5, 6.5) {$0$}; 
    \node[] at (\offs+0.5, 5.5) {$0$}; 
    \node[] at (\offs+0.5, 4.5) {$\vdots$}; 
    \node[] at (\offs+0.5, 3.5) {$0$}; 
    \node[] at (\offs+0.5, 2.5) {$q_{r \Plus 1, 1}$}; 
    \node[] at (\offs+0.5, 1.5) {$\vdots$}; 
    \node[] at (\offs+0.5, 0.5) {$q_{d, 1}$}; 

    \node[] at (\offs+1.5, 8.5) {$q_{1,2}$};
    \node[] at (\offs+1.5, 7.5) {$q_{2,2}$};
    \node[] at (\offs+1.5, 6.5) {$0$};
    \node[] at (\offs+1.5, 5.5) {$0$};
    \node[] at (\offs+1.5, 4.5) {$\vdots$}; 
    \node[] at (\offs+1.5, 3.5) {$0$}; 
    \node[] at (\offs+1.5, 2.5) {$q_{r \Plus 1, 2}$}; 
    \node[] at (\offs+1.5, 1.5) {$\vdots$}; 
    \node[] at (\offs+1.5, 0.5) {$q_{d, 2}$}; 

    \node[] at (\offs+2.5, 8.5) {$q_{1,3}$};
    \node[] at (\offs+2.5, 7.5) {$q_{2,3}$};
    \node[] at (\offs+2.5, 6.5) {$q_{3,3}$};
    \node[] at (\offs+2.5, 5.5) {$0$}; 
    \node[] at (\offs+2.5, 4.5) {$\vdots$}; 
    \node[] at (\offs+2.5, 3.5) {$0$};
    \node[] at (\offs+2.5, 2.5) {$q_{r \Plus 1, 3}$}; 
    \node[] at (\offs+2.5, 1.5) {$\vdots$}; 
    \node[] at (\offs+2.5, 0.5) {$q_{d, 3}$}; 

    \node[] at (\offs+3.5, 8.5) {$\hdots$};
    \node[] at (\offs+3.5, 7.5) {$\hdots$};
    \node[] at (\offs+3.5, 6.5) {$\hdots$};
    \node[] at (\offs+3.5, 5.5) {$\hdots$};
    \node[] at (\offs+3.5, 4.5) {$\ddots$}; 
    \node[] at (\offs+3.5, 3.5) {$\hdots$}; 
    \node[] at (\offs+3.5, 2.5) {$\hdots$}; 
    \node[] at (\offs+3.5, 1.5) {$\ddots$}; 
    \node[] at (\offs+3.5, 0.5) {$\hdots$}; 

    \node[] at (\offs+4.5, 8.5) {$q_{1,r \Minus 1}$};
    \node[] at (\offs+4.5, 7.5) {$q_{2,r \Minus 1}$};
    \node[] at (\offs+4.5, 6.5) {$q_{3,r \Minus 1}$};
    \node[] at (\offs+4.5, 5.5) {$q_{4,r \Minus 1}$};
    \node[] at (\offs+4.5, 4.5) {$\vdots$}; 
    \node[] at (\offs+4.5, 3.5) {$0$}; 
    \node[] at (\offs+4.5, 2.5) {$q_{r \Plus 1, r \Minus 1}$}; 
    \node[] at (\offs+4.5, 1.5) {$\vdots$}; 
    \node[] at (\offs+4.5, 0.5) {$q_{d, r \Minus 1}$}; 

    \node[] at (\offs+5.5, 8.5) {$q_{1,r}$};
    \node[] at (\offs+5.5, 7.5) {$q_{2,r}$};
    \node[] at (\offs+5.5, 6.5) {$q_{3,r}$};
    \node[] at (\offs+5.5, 5.5) {$q_{4,r}$};
    \node[] at (\offs+5.5, 4.5) {$\vdots$}; 
    \node[] at (\offs+5.5, 3.5) {$q_{r,r}$};
    \node[] at (\offs+5.5, 2.5) {$q_{r \Plus 1, r}$}; 
    \node[] at (\offs+5.5, 1.5) {$\vdots$}; 
    \node[] at (\offs+5.5, 0.5) {$q_{d, r}$}; 

\end{tikzpicture}%
%
}
\caption{
Visualization of Householder parameterizations of orthonormal frames as per the LAPACK convention: (a) Full parameterization, (b) Reduced parameterization. Shaded areas with $h_{i,j}$ values represent parameters of reflectors; $q_{i,j}$ values represent orthonormal frame elements, affected by the corresponding parameterization.
}
\label{fig:hh}
\end{figure*}
To construct the mapping $\phi$ from the parameter space to orthonormal frames, we utilize a sequence of Householder reflections~\citep{shepard2015representation}.
It is known that any matrix $A\in\mathbb{R}^{d\times r}$ can be represented using QR-decomposition $A = QR$, where $Q\in \mathrm{St}(d, r)$ and $R\in\mathbb{R}^{r\times r}$ is upper-triangular.
The matrix $Q$ can be given as a product of Householder reflections:
\begin{equation}\label{eq:orgqr}
    Q = H^{(1)} H^{(2)} \dots H^{(r)} I_{d\times r},
\end{equation}
where $I_{d\times r}$ is a truncated identity matrix of size $d\times r$, and the Householder reflector $H^{(i)}$ is written as $H^{(i)} = I_d - 2 u^{(i)} u^{(i)\top}$ for some $u^{(i)}\in\mathbb{R}^{d}$: $\|u^{(i)}\|_2 = 1$ and $u^{(i)}_\alpha = 0$, $\alpha=1,\dots, i-1$.

A QR decomposition of $U\in \mathrm{St}(d, r)$ results in a diagonal matrix $R$ with $R_{ii} = \{-1,1\}, i=1,\dots,r$.
To ensure the uniqueness and differentiability of SVDP, one has to choose $R_{ii}$ carefully; it affects the numerical stability of the parameterization of certain regions of $\mathrm{St}(d,r)$.
As a result, $U=QR$, and hence it is given by the matrices $H^{(i)}$.
The number of \textit{degrees of freedom} ($\mathrm{DOF}$) to represent $H^{(i)}$ is $(d-i)$ since we only need to store nonzero entries of $u^{(i)}$, and there is an additional requirement $\|u^{(i)}\|_2 = 1$. 
The total number of parameters to represent all the $H^{(i)}$, $i=1,\dots, r$ is
\[
   \mathrm{DOF}(U) = \sum_{i=1}^r (d-i) = dr - \frac{r(r+1)}{2},
\]
which coincides with the dimensionality of $\mathrm{St}(d, r)$.

We use the LAPACK convention for parameter layout in a matrix with columns $h^{(i)}, i=1,...,r$ (Fig. \ref{fig:hh}\hyperref[fig:hh]{a}) and obtain $u^{(i)}=h^{(i)} / \|h^{(i)}\|_2$. 

Once both $U\in\mathrm{St}(d_{\dout}, r)$ and $V\in\mathrm{St}(d_{\din}, r)$ are parameterized as is described above, SVDP spans the whole manifold of weight matrices of ranks not higher than $r$.
Thus, the total number of degrees of freedom required to parameterize the matrix $W \in \mathbb{R}^{d_{\dout} \times d_{\din}}$ using SVDP with rank $r$ adds up from the numbers of parameters required to parameterize orthonormal frames $U\in\mathbb{R}^{d_{\dout}\times r}$, $V\in\mathbb{R}^{d_{\din} \times r}$ and $r$ singular values:
\[
    \mathrm{DOF}(W) = r (d_{\dout} + d_{\din}) - r^2.
\]

\subsection{The Case of Identity Spectrum}
\label{sec:identityspectrum}

When all $r$ singular values are fixed to $1$ ($\Sigma=I_r$), independent parameterizations of $U$ and $V$ lead to redundancy in $W$.
This is due to the fact that for any orthogonal matrix $Q \in \mathbb{R}^{r\times r}$, the following holds:
\begin{equation}
\label{lbl:eq:svd_redundancy}
    W = UV^\top =UQ^\top QV^\top= (UQ) (VQ)^\top,
\end{equation}
which leads to a new $\widetilde U = UQ\in \mathrm{St}(d_{\dout}, r)$ and $\widetilde V = VQ\in \mathrm{St}(d_{\din}, r)$ that produce the same $W$.

To eliminate this redundancy of parameters, we impose additional constraints on either $U$ or $V$ (we choose $U$ for concreteness).
We follow the Grassmann manifold parameterization~\citep{shepard2015representation} and require the leading $r\times r$ sub-matrix of $U\in\mathrm{St}(d_{\dout}, r)$ to be upper triangular.
We denote the subset of all such matrices by $\mathrm{St}_{\mathsf{U}}(d_{\dout}, r) \subset \mathrm{St}(d_{\dout}, r)$ (subscript $\mathsf{U}$ for ``upper'').

To parameterize a matrix $U \in \mathrm{St}_{\mathsf{U}}(d_{\dout}, r)$, we propose a reduced form of the Householder parameterization (Fig.~\ref{fig:hh}\hyperref[fig:hh]{b}).
It differs from the full parameterization by setting the entries $h_j^{(i)}, i+1 \leq j < r$ to zero.
This saves us $r(r-1)/2$ parameters to store the vectors $h^{(i)}$, $i=1,\dots,r$, leading to $(d_{\dout}r - r^2)$ effective parameters.

Thus the total number of independent parameters required to parameterize $W$ with $U\in \mathrm{St}_{\mathsf{U}}(d_{\dout}, r)$, $V\in \mathrm{St}(d_{\din}, r)$, and $\Sigma=I_r$ becomes
\[
    \mathrm{DOF}(W) = r(d_{\dout} + d_{\din}) - \frac{r(3r+1)}{2}.
\]
Apart from a smaller parameter footprint, redundancy removal may also benefit the optimization landscape, as the redundant parameters introduce plateau regions.

\subsection{STTP}
\label{sec:param_stiefel_tt}

We assume that the dimensions of the weight matrix $W\in\mathbb{R}^{d_{\dout}\times d_{\din}}$ factorize (see discussion in Sec.~\ref{lbl:sec:architecturedesign}): $d_{\dout}=n_1^{\dout}\cdots n_{D_\dout}^{\dout}$, $d_{\din}=n_1^{\din}\dots n_{D_\din}^{\din}$, for example, prime factors with repetition. 
Thus we can consider parameterizing the weight matrix $W$ tensorized into a tensor $\widetilde W$ with factored dimensions ${n_1^{\dout}\times \dots \times  n_{D_\dout}^{\dout} \times n_{D_\din}^{\din}\times \dots \times n_1^{\din}}$. 
Upon obtaining $\widetilde W$ from the underlying parameters, the matrix structure of $W$ can be recovered through matricization.

As previously discussed, simply parameterizing $\widetilde W$ as a TT decomposition with unconstrained parameterizations of TT-cores~\eqref{eq:ttprelim} as done in the prior art does not specify the spectral properties of $W$. 
However, since the TT decomposition is inherently redundant, $\widetilde W$ can have multiple equivalent TT decompositions, including the one shown in Fig.~\ref{fig:tt_network}. 
Here matricizations $\mathcal{M}\colon \mathbb{R}^{a \times b \times c}\to \mathbb{R}^{ab \times c}$ of TT-cores $\mathcal{U}^{(i)}$ and $\mathcal{V}^{(j)}$ are\footnote{
  Similar to how matrices $U$ and $V$ are treated identically in Sec.~\ref{sec:param_stiefel_full}, so are TT-cores of $U$ and $V$; however, the transposed $V$ in \eqref{eq:svd} leads to the transposed $\mathcal{M}(\mathcal{V}^{(k)})$. 
  Therefore, TT-cores of $V^\top$ are denoted as $\mathcal{V}^{(k)\top}$ in Fig.~\ref{fig:tt_network}.
}
orthonormal frames, and $\Sigma$ is a matrix of singular values of the weight matrix $W$~\citep{holtz2012manifolds}.

Let us show that this TT decomposition can be reduced to the SVD form~\eqref{eq:svd}, with $U$ and $V$ being orthonormal frames. 
The elements of the matrix $U\in\mathbb{R}^{n_1\cdots n_D \times r}$ are products of TT-cores $\mathcal{U}^{(k)}$ to the left of $\Sigma$ in Fig.~\ref{fig:tt_network}:
\begin{equation}
\label{eq:tt}
    U_{\overline{i_1\dots i_D}, \alpha}
    = \!\!\!\!\!\! \sum_{\beta_0,\ldots,\beta_D=1}^{R_0,..,R_D} \!\!\!
    \mathcal{U}_{\beta_0, i_1,\beta_1}^{(1)} \!\! \cdot
    \mathcal{U}_{\beta_1,i_2,\beta_2}^{(2)} \!\! \dots \,
    \mathcal{U}_{\beta_{D\text{-}1}, i_{D}, \alpha}^{(D)},
\end{equation}
where $\overline{i_1\dots i_D}$ is computed as in~\eqref{eq:order}, $\alpha \in [1,r]$.
The next proposition illustrates that our choice of TT-cores $\mathcal{U}^{(k)}$ leads to $U\in\mathrm{St}(d,r)$.
For convenience, we use the notation $U = \mathcal{T}(\mathcal{U}^{(1)},\dots,\mathcal{U}^{(D)})$ as a shorthand to~\eqref{eq:tt}.

\begin{prop} 
\label{prop:ttorth}
Let the matricizations $\mathcal{M} (\mathcal{U}^{(k)})\in \mathbb{R}^{R_{k-1} n_k \times R_k}$ of the TT-cores $\mathcal{U}^{(k)}\in \mathbb{R}^{R_{k-1} \times n_k \times  R_{k}}$ satisfy $\mathcal{M} (\mathcal{U}^{(k)}) \in \mathrm{St}(R_{k-1}n_k, R_k)$, $k=1,\dots,D$.
Then $\mathcal{T}(\mathcal{U}^{(1)},\dots,\mathcal{U}^{(D)})\in\mathrm{St}(n_1\dots n_D, r)$. 
\end{prop}

\textit{Proof.} Follows from~\citet[Lemma 3.1]{oseledets2011tensor}; see the proof in Sec.~\ref{sec:proof1} for completeness. \hfill$\square$

Proposition~\ref{prop:ttorth} gives us a framework to perform parameterization of TT-cores, leading to parameterizations of $U$ and $V$, and in the end, $W$ with a given spectrum.
It follows that we can parameterize the matricized TT-cores $\mathcal{M} (\mathcal{U}^{(k)})\in\mathrm{St}(R_{k-1}n_k, R_k)$ using the procedure described in Sec.~\ref{sec:param_stiefel_full}.
Nevertheless, the following proposition suggests that parameterizing each $\mathcal{M}(\mathcal{U}^{(k)})$ simply as an element of the Stiefel manifold leads to over-parameterization similar to~\eqref{lbl:eq:svd_redundancy}, which is a direct consequence of TT decomposition non-uniqueness.

\begin{prop}
\label{prop:ttredundancy}
Let $Q_k\in\mathbb{R}^{R_k\times R_k}$ be orthogonal for $k=1,\dots,D{-}1$, $Q_0=1$, $Q_D=I_r$.
We also assume that $\mathcal{U}^{(k)}\in \mathbb{R}^{R_{k-1} \times n_k \times R_k}$, $k=1,\dots, D$ are such that $\mathcal{M} (\mathcal{U}^{(k)})\in\mathrm{St}(R_{k-1}n_k, R_k)$. 
We define $\widetilde{\mathcal{U}}^{(k)}\in\mathbb{R}^{R_{k-1}\times n_k \times R_k}$: $\widetilde{\mathcal{U}}_{:,i_k,:}^{(k)} = Q_{k-1}^\top\mathcal{U}_{:,i_k,:}^{(k)} Q_k$
(where $\mathcal{U}_{:,i_k,:}^{(k)} \in \mathbb{R}^{R_{k-1} \times R_k}$).
Then 
\begin{equation*}
\begin{split}
  & \mathcal{M}(\widetilde{\mathcal{U}}^{(k)}) \in \mathrm{St}(R_{k-1}n_k, R_k), \quad k=1,\dots,D, \quad \mathrm{and} \\
  & \mathcal{T}\left(\mathcal{U}^{(1)}, \dots,\mathcal{U}^{(D)}\right) =  \mathcal{T}\left(\widetilde{\mathcal{U}}^{(1)}, \dots, \widetilde{\mathcal{U}}^{(D)}\right).
\end{split}
\end{equation*}
\end{prop}

\textit{Proof.} See a complete proof in Sec.~\ref{sec:proof2}. \hfill$\square$

To avoid over-parameterization, we reuse the approach from Sec.~\ref{sec:identityspectrum} and require all TT-cores except for the two adjacent to $\Sigma$ in Fig.~\ref{fig:tt_network} to have reduced parameterizations: $\mathcal{M} (\mathcal{U}^{(k)})\in\mathrm{St}_{\mathsf{U}}(R_{k-1}n_k, R_k)$, $k=1,\dots,D-1$ and $\mathcal{M} (\mathcal{U}^{(D)})\in\mathrm{St}(R_{D{-}1}n_D, r)$.
For the edge case of identity spectrum, we additionally require $\mathcal{M}(\mathcal{U}^{(D)})\in\mathrm{St_{\mathsf{U}}}(R_{D-1}n_D, r)$ just for the last TT-core of $U$ (but not for the last TT-core of $V$).
The algorithm to enforce such parameterizations is described in Sec.~\ref{sec:param_stiefel_full}.

The total number of independent parameters required to parameterize $W$ with STTP without redundancy is:
\begin{equation} 
\label{eq:ttdof}
\begin{split}
    &\mathrm{DOF}(W) = 
     \negthickspace\negthickspace\negthickspace
     \sum_{k=1}^{D_{\dout} + D_{\din}} 
     \negthickspace\negthickspace\negthickspace 
     R_{k{-}1} n_k R_k 
     - 
     \negthickspace\negthickspace\negthickspace\negthickspace\negthickspace
     \sum_{k=1}^{D_{\dout} + D_{\din}-1} 
     \negthickspace\negthickspace\negthickspace\negthickspace
     R_k^2, \ \ \ \ \mathrm{where}
     \\
    &R=(1, R_1^{\dout}, \ldots, R_{D_{\dout}-1}^{\dout}, r
    , R_{D_{\din}-1}^{\din}, \ldots, R_1^{\din}, 1)\footnotemark,
     \\
    &n=(n_1^{\dout}, \ldots, n_{D_{\dout}}^{\dout}, n_{D_{\din}}^{\din}, \ldots, n_1^{\din})
\end{split}
\end{equation}
\footnotetext{
TT-rank of a tensor diagram Fig.~\ref{fig:tt_network} made compatible with the form of TT decomposition introduced in~\eqref{eq:ttprelim} and Fig.~\ref{fig:ttnetwork:simple} (consisting only of TT-cores) by contracting the matrix $\Sigma$ into either left or right adjacent TT-core.
}%
are the TT-rank and dimensions of $\widetilde W$ respectively.
Notably, $\mathrm{DOF}(W)$ with learned spectrum matches the dimensionality of the fixed TT-rank $R$ tensor manifold given in~\citep{holtz2012manifolds} (see derivation in Sec.~\ref{sec:sttp:dof}). 

An edge case of STTP happens with the values of TT-rank $R$ (excluding $r$ in the middle) set to $R_k^{\mathrm{max}}$ subject to~\eqref{eq:ttrankcapprelim}: such parameterization spans the same manifold of rank-$r$ matrices $W$ as SVDP. 
A careful inspection of this edge case reveals that it is an SVDP in disguise: all TT-cores except for the two adjacent to $\Sigma$ have square matricizations. 
Given that $\forall p$ $\mathrm{St}_\mathsf{U}(p,p)$ does not require any learned parameters (Fig.~\ref{fig:hh}\hyperref[fig:hh]{b}), all of them are concentrated in $\Sigma$ and the two adjacent TT-cores with matricizations of sizes $d_\dout \times r$ and $d_\din \times r$.

The degree of compression of $W$ is defined by the TT-rank $R$. In practice, we treat $r$ as the only hyperparameter and compute TT-rank values as $R_k = \min (r, R_k^{\mathrm{max}})$ using~\eqref{eq:ttrankcapprelim}.
As such, $\mathrm{DOF}(W) = \mathcal{O}(r^2 \log(d_{\dout} d_{\din}))$; the number of parameters is logarithmic in the size of $W$. 

\subsection{Spectral Constraints}
\label{sec:spectralconstraints}

We consider two distinct cases: the identity spectrum (Sec.~\ref{sec:param_stiefel_full}) and learned parameterization of the diagonal matrix $\Sigma$, with optional regularization. 
As was previously shown, the former case results in a more compact (also more restricted) parameterization.

The learned singular values $\Sigma$ are parameterized with a vector $S \in \mathbb{R}^r$.
To implement a Lipschitz-1 constraint~\eqref{eq:lipsch}, we initialize $S=I_r$ and compute $\Sigma$ to keep all singular values constrained in the $[-1, 1]$ range:
\[
    \Sigma = \diag(S / \|S\|_\infty)
\]
Furthermore, we investigate whether an additional regularization term associated with $\Sigma$ helps to learn a better model. 
To this end, we explore the \mbox{D-optimal} regularizer~\eqref{eq:doptreg} as in~\cite{jiang2018computation}, which penalizes the learned singular values of small magnitude:
\begin{equation}
\label{eq:doptreg}
    \mathcal{R}(\Sigma) = -\sum_{i=1} \log |\Sigma_{i}|.
\end{equation}
As will be discussed in the experiments section, the spectra of neural network layers are crucial to the stability of optimization and good model performance. 
A near-zero element in $\Sigma$ effectively reduces the rank of the whole weight matrix $W$. 
Therefore, embedding spectral properties through regularization or identity spectrum is a more versatile approach than enforcing only the Lipschitz-1 constraint on the model.

\section{Experiments}
\label{sec:exp}

\begin{figure}[t]
\centering
\includegraphics[width=0.49\linewidth]{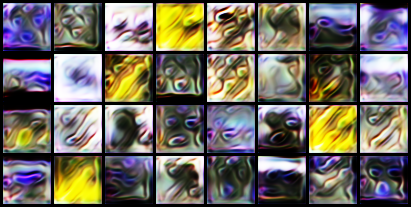}
\hfill
\includegraphics[width=0.49\linewidth]{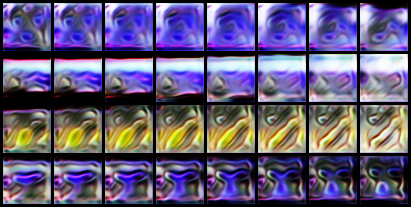}\\[0.3em]
\includegraphics[width=0.49\linewidth]{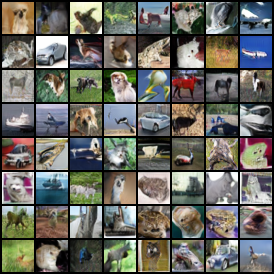}
\hfill
\includegraphics[width=0.49\linewidth]{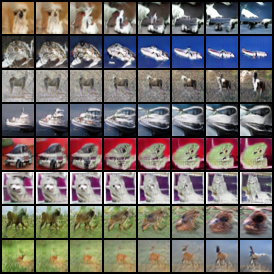}
\caption{
Samples from two trained SNGAN~\citep{miyato2018spectral} generators (top, bottom). 
Top: spectral collapse during training leads to degenerate samples at test time. 
Bottom: diverse samples from the generator. 
Left: random samples. 
Right: latent code interpolation between left-most and right-most random samples.
}
\label{fig:gan:results}
\end{figure}

\begin{figure}[t]
\centering
\includegraphics[width=\linewidth]{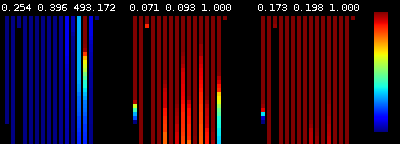}
\caption[]{
Visualization of singular values (SV) of SNGAN-48 discriminators ($\mathcal{D}$), trained with SN~\citep{miyato2018spectral} (left), SVDP-L (middle), and SVDP-R (right). 
Each column represents one layer of $\mathcal{D}$ in the feed-forward order. 
Column height represents the rank of a particular layer. 
For example, the first layer is a Conv2D with 64 outputs, 3 inputs, and $3\times3$ kernel; hence its rank is 27. 
Only the top 32 SVs are shown. 
Three annotations at the top correspond to the minimum SV in $\mathcal{D}$, the minimum SV in the observed slice of 32 SVs (blue), and the maximum SV in $\mathcal{D}$ (red). 
SN spectrum grows unconstrained and unbalanced, which causes a spectral collapse in more restricted configurations, such as the one from Table~\ref{tbl:results:gan:reduced}.
}
\label{fig:svs}
\end{figure}

\begin{table*}[ht!]
\centering
\caption{
Results of training the SNGAN model with an unparameterized generator ($\mathcal{G}$) and a reduced discriminator ($\mathcal{D}$) on two image datasets. 
Methods: SN -- Spectral Normalization~\citep{miyato2018spectral}, SR -- Spectral Regularization~\citep{liu2019spectral}, SVDP-C, STTP-C -- the proposed methods with the identity spectrum and rank $r=64$.
Metrics: $\mathcal{Z}$ -- the ratio~\eqref{eq:compression} of counts of the learned discriminator parameters relative to SN (lower is better), IS -- Inception Score (higher is better), FID -- Fr\'echet Inception Distance (lower is better), and KID -- Kernel Inception Distance (lower is better).
The reduction of discriminator features causes a spectral collapse in layer weight matrices in the original (SN) setting. 
Both SR and our methods prevent spectral collapse.
}
\label{tbl:results:gan:reduced}
\begin{tabular}{@{}l c cccc c cccc@{}}

\toprule 

Dataset && 
\multicolumn{4}{c}{CIFAR10} && 
\multicolumn{4}{c}{STL10\,--\,48} \\

\cmidrule{3-6} \cmidrule{8-11} 

Metric &&
$\mathcal{Z}$ $\downarrow$ & 
IS $\uparrow$ & 
FID $\downarrow$ & 
KID$\times100$ $\downarrow$ &&
$\mathcal{Z}$ $\downarrow$ & 
IS $\uparrow$ & 
FID $\downarrow$ & 
KID$\times100$ $\downarrow$ \\ 

\midrule 

SN &&
100 & 
  5.48\rpm 0.44 & 
 55.17\rpm 3.65 & 
  3.84\rpm 0.15 && 
100 & 
  2.56\rpm 0.24 & 
251.6\rpm 26.4 & 
 28.78\rpm 5.28 \\

SR && 
100 & 
7.17\rpm 0.06 & 
27.24\rpm 0.80 & 
1.95\rpm 0.06 && 
100 & 
3.77\rpm 0.32 & 
193.9\rpm 11.5 & 
19.55\rpm 1.36 \\

\cmidrule{3-6} \cmidrule{8-11} 

SVDP-C && 
93.1 & 
7.42\rpm 0.20 & 
23.76\rpm 2.12 & 
1.77\rpm 0.24 && 
89.4 & 
3.91\rpm 0.36 & 
204.3\rpm 16.9 & 
21.48\rpm 1.77 \\

STTP-C && 
87.7 & 
7.38\rpm 0.08 & 
24.45\rpm 0.48 & 
1.76\rpm 0.08 && 
74.1 & 
4.35\rpm 0.16 & 
190.6\rpm 13.7 & 
19.59\rpm 1.96 \\

\bottomrule
\end{tabular}
\end{table*}

\begin{table*}[ht!]
\centering
\caption{
Results of training the SNGAN model with an unparameterized generator ($\mathcal{G}$) and the full discriminator ($\mathcal{D}$) on two image datasets.
Added methods: ``-L" -- learned spectrum, ``-R" -- learned and regularized spectrum~\citep{jiang2018computation}. Metrics: see Table.~\ref{tbl:results:gan:reduced}. All models parameterized with SVDP and STTP use rank $r=64$. 
Both SVDP and STTP result in smaller models with comparable performance.
}
\label{tbl:results:gan:original}
\begin{tabular}{@{}l c cccc c cccc@{}}

\toprule 

Dataset && 
\multicolumn{4}{c}{CIFAR10} && 
\multicolumn{4}{c}{STL10\,--\,48} \\

\cmidrule{3-6} \cmidrule{8-11} 

Metric &&
$\mathcal{Z}$ $\downarrow$ & 
IS $\uparrow$ & 
FID $\downarrow$ & 
KID$\times100$ $\downarrow$ &&
$\mathcal{Z}$ $\downarrow$ & 
IS $\uparrow$ & 
FID $\downarrow$ & 
KID$\times100$ $\downarrow$ \\ 

\midrule 

SN &&
100 & 
7.99\rpm 0.02 & 
18.38\rpm 0.30 & 
1.25\rpm 0.04 &&
100 & 
8.38\rpm 0.11 & 
99.26\rpm 0.58 & 
10.56\rpm 0.01 \\

SR && 
100 & 
8.02\rpm 0.05 & 
16.89\rpm 0.23 & 
1.17\rpm 0.02 && 
100 & 
8.85\rpm 0.11 & 
93.14\rpm 0.56 & 
09.59\rpm 0.11 \\

\cmidrule{3-6} \cmidrule{8-11} 

SVDP-C && 
51.7 & 
7.97\rpm 0.08 & 
17.53\rpm 0.63 & 
1.23\rpm 0.05 && 
14.2 & 
8.49\rpm 0.12 & 
98.50\rpm 1.60 & 
10.40\rpm 0.22 \\

STTP-C && 
16.7 & 
7.85\rpm 0.01 & 
18.93\rpm 0.54 & 
1.37\rpm 0.01 &&
3.24 & 
8.59\rpm 0.11 & 
96.91\rpm 0.89 & 
10.22\rpm 0.07 \\

\cmidrule{3-6} \cmidrule{8-11} 

SVDP-L && 
53.3 & 
7.97\rpm 0.06 & 
17.18\rpm 1.30 & 
1.24\rpm 0.15 && 
14.5 & 
8.71\rpm 0.04 & 
96.03\rpm 1.53 & 
10.18\rpm 0.27 \\

STTP-L && 
18.3 & 
7.96\rpm 0.02 & 
18.00\rpm 0.56 & 
1.24\rpm 0.03 &&
3.51 & 
8.64\rpm 0.07 & 
96.04\rpm 1.57 & 
10.12\rpm 0.14 \\

\cmidrule{3-6} \cmidrule{8-11} 

SVDP-R && 
53.3 & 
8.02\rpm 0.07 & 
17.17\rpm 0.37 & 
1.23\rpm 0.03 && 
14.5 & 
8.69\rpm 0.07 & 
95.75\rpm 0.84 & 
10.14\rpm 0.18 \\

STTP-R && 
18.3 & 
7.96\rpm 0.01 & 
18.43\rpm 0.75 & 
1.32\rpm 0.06 &&
3.51 & 
8.67\rpm 0.10 & 
97.12\rpm 0.68 & 
10.29\rpm 0.22 \\

\bottomrule
\end{tabular}
\end{table*}

We evaluate STTP in the Generative Adversarial Networks (GANs) training setting, which is unstable due to its non-convex non-concave optimization objective. 
The \textit{Mode Collapse} problem goes back to~\cite{goodfellow2014generative} where it manifested in the generator producing data samples of limited variety (Fig.~\ref{fig:gan:results}, top). 

Recent analysis~\citep{che2016mode,Arjovsky2017TowardsPM,gulrajani2017improved} suggests that the family of discriminator functions plays a key role in the training dynamics (i.e., training stability) and the quality of the generator parameters updates.
Others~\citep{miyato2018spectral,jiang2018computation} tied up mode collapse in the generator with Lipschitz continuity of the discriminator, and most recently, \cite{liu2019spectral} associated mode collapse with the \textit{Spectral Collapse}. 
The latter is a condition of simultaneous growth of the spectral norm and drop of the \textit{stable rank}~\citep{sanyal2019stable} of weight matrices in the discriminator layers. 

\subsection{Setup}

We follow closely the unconditional image generation setup of SNGAN from~\cite{miyato2018spectral}; in particular, we use a residual generator $\mathcal{G}$ and discriminator $\mathcal{D}$ for image sizes $32\times32$ and $48\times48$ and the hinge loss objective $\mathrm{min_\mathcal{G}\, max_\mathcal{D}\, }V\mathrm{(\mathcal{G}, \mathcal{D})}$. 

Experiments with $32\times32$ size are conducted with the CIFAR-10~\citep{krizhevsky2009learning} dataset, consisting of 50K images of 10 classes. 
For images of $48\times48$ size, we utilize the unlabeled split of the STL-10~\citep{coates2011analysis} dataset, consisting of 100K images.

All experiments are trained on a single 11GB GPU for 100K generator updates, 5 discriminator updates per one generator update, batch size 64, Adam optimizer with betas $[0, 0.9]$, and the learning rate $2e{\mathrm{-}}4$, with linear decay of the learning rate to zero towards the end of the training. 
All code is implemented in \texttt{PyTorch}~\citep{pytorch} for consistency of comparisons. For spectral normalization, we use the standard \texttt{torch.nn.utils.spectral\_norm} with one power iteration per update.
To transform the learned parameters of SVDP and STTP into weight matrices, we employ \texttt{torch-householder}~\citep{obukhov2021torchhouseholder} to compute~\eqref{eq:orgqr} and \texttt{opt\_einsum}~\citep{Smith2018opteinsum} to contract tensor diagrams in Fig.~\ref{fig:networks}. See details in Sec.~\ref{sec:compshortcuts},~\ref{lbl:batchedhh}.

For assessing the generated results, we use the \textit{Inception Score} (IS)~\citep{salimans2016improved} (higher is better) and the \textit{Fr\'echet Inception Distance} (FID)~\citep{heusel2017gans} (lower is better). 
Both IS and FID are known to correlate with the human perception of sample quality, however, IS is known to fluctuate inadequately in the presence of synthetic artifacts. 
We also report the \textit{Kernel Inception Distance} (KID)~\citep{binkowski2018demystifying} (lower is better), as it was shown to have no bias, thus making its values comparable across a wider range of evaluation protocols (e.g., different sample and subset sizes).
During the evaluation, we sample 50000 images from the generator and compute IS with ten splits, FID with all samples, KID with 100 subsets each 1000 samples. 
For all metrics, we report the mean and two standard deviations (68\%) confidence interval over three runs with different seeds.
Evaluation is performed with \texttt{torch-fidelity}~\citep{obukhov2020torchfidelity}, which is shown to be consistent with reference implementations.

For each reparameterized model, we calculate the compression ratio using the following formula:
\begin{equation}
\label{eq:compression}
    \mathcal{Z}\left(\mathcal{F}\right) = 100 \times \frac{
        \sum_{\ell=1}^L \mathrm{DOF}(W_{\ell}) + C_{\ell}
    }{
        \sum_{\ell=1}^L \mathrm{numel}(W_{\ell}) + C_{\ell}
    },
\end{equation}
where $\mathrm{DOF}(W_{\ell})$ denotes the number of degrees of freedom of the weight matrix $W_{\ell}$ with respect to the chosen parameterization, $\mathrm{numel}(W_{\ell})$ is the number of elements in the weight matrix $W_{\ell}$, and $C_{\ell}$ is the number of parameters not subject to reparameterization, such as bias terms $b_{\ell}$ and parameters of batch norms. 

\subsection{Effect of Spectral Constraints}

We compare with two methods: spectral normalization (SN)~\citep{miyato2018spectral} and spectral regularization (SR)~\citep{liu2019spectral} applied in place of SN.
First, we want to verify that the proposed parameterization prevents spectral collapse when it is known to happen under SN. 
We use the same strategy as~\cite{liu2019spectral}, who showed that reducing the number of channels in all layers of the discriminator leads to both spectral and mode collapses. 
We reduce the number of channels in the discriminators of SNGAN-32 and SNGAN-48 by $4\times$ (128 to 32) and $16\times$ (1024 to 64), respectively. 
Such reduction limits the discriminator's ability to provide good updates to the generator. 
We use the identity spectrum to leave out the spectrum factor of variation. 
All models are trained with a rank $r\leq64$. 

The first observation is that SR consistently prevents spectral collapse. 
We consider SR as an improved version of SN, which performs normalization and maintains weight matrices' stable ranks. 
It is worth noting that SR could be seen as a full-rank method, as it applies SVD on each weight matrix every training step. 
Therefore, we group SN and SR in Table~\ref{tbl:results:gan:reduced} as baselines. 
We aim to demonstrate that our low-rank method is better than SN and around or better than SR. 
Results of the reduced discriminator experiments confirm that both SVDP and STTP consistently outperform SN even with the identity spectrum. 

In the second set of experiments, we use the original networks with unaltered channels of the discriminators; hence both SN and SR perform well with no spectral collapse. 
We compare various strategies of spectrum control in Table~\ref{tbl:results:gan:original}.
As before, our method outperforms SN even with the identity spectrum (-C suffix). 
Letting singular values loose (-L suffix) improves our method over the identity spectrum, but the best performance is achieved with the D-optimal regularizer (-R suffix). 
We conjecture that it allows the optimizer to take shortcuts when traversing the manifolds of $W$; however, such a regularizer makes singular values end up close to 1, as confirmed by Fig.~\ref{fig:svs}. 
All models from Table~\ref{tbl:results:gan:original} produce visually appealing results (Fig.~\ref{fig:gan:results}, bottom).

\subsection{Rank Utilization Study}
\label{lbl:svdvstt}

\begin{table*}[t!]
\centering
\caption{
Results of varying the rank hyperparameter ($r$) in GAN modules on CIFAR-10 image generation with SVDP and STTP separately applied to the generator ($\mathcal{G}$) and the discriminator ($\mathcal{D}$). 
Experiments with parameterized $\mathcal{D}$ (left) use unconstrained $\mathcal{G}$. Additionally, experiments with parameterized $\mathcal{G}$ (right) use SVDP of $\mathcal{D}$ with rank $r=64$. 
All experiments use the \mbox{D-optimal} regularizer. 
Performance of STTP with $2{-}3\times$ rank is similar to that of SVDP, but with higher compression $\mathcal{Z}$~\eqref{eq:compression} in terms of the number of parameters (DOF).
}
\label{tbl:ranks}
\begin{tabular}{@{}cc c cccc c cccc@{}}

\toprule 

& 
Scope && 
\multicolumn{4}{c}{Discriminator ($\mathcal{D}$)} && 
\multicolumn{4}{c}{Generator ($\mathcal{G}$)} \\

\cmidrule{4-7} \cmidrule{9-12} 

&
Metric &&
$\mathcal{Z}$ $\downarrow$ &
IS $\uparrow$ & 
FID $\downarrow$ & 
KID$\times100$ $\downarrow$ &
&
$\mathcal{Z}$ $\downarrow$ & 
IS $\uparrow$ & 
FID $\downarrow$ & 
KID$\times100$ $\downarrow$ \\ 

\midrule 

\multirow{2}{*}{\rotatebox{90}{\tiny{SVDP}}} &

$r{=}32$ &&
 27.71 & 
 7.94\rpm 0.08 & 
 18.33\rpm 0.44 &
 1.36\rpm 0.01 &
 & 
 25.14 & 
 7.82\rpm 0.07 & 
 19.46\rpm 0.64 &
 1.39\rpm 0.03 \\

&
$r{=}64$ &&
 53.36 & 
 8.02\rpm 0.07 & 
 17.17\rpm 0.37 &
 1.23\rpm 0.03 &
 & 
 37.13 & 
 8.05\rpm 0.07 & 
 17.09\rpm 0.11 &
 1.23\rpm 0.05 \\

\midrule 

\multirow{2}{*}{\rotatebox{90}{\tiny{STTP}}} &

$r{=}32$ &&
 06.44 & 
 7.51\rpm 0.11 & 
 25.16\rpm 0.57 &
 1.88\rpm 0.10 &
 & 
 14.61 & 
 7.07\rpm 0.09 & 
 31.32\rpm 2.09 &
 2.29\rpm 0.13 \\

&
$r{=}64$ &&
 18.33 & 
 7.91\rpm 0.06 & 
 17.89\rpm 0.65 &
 1.28\rpm 0.01 &
 & 
 18.82 & 
 7.71\rpm 0.03 & 
 22.06\rpm 0.05 &
 1.61\rpm 0.02 \\

\bottomrule
\end{tabular}
\end{table*}

For simplicity, all our experiments use the same maximum rank hyperparameter (denoted as $r$) in all layers. 
This design does not limit the model's ability to learn low-rank projections when learning the spectrum (e.g., with D-optimal regularizer), as any rank-$\rho$ projection ($\rho < r$) can be achieved by setting a subset of $r-\rho$ singular values to zero. 
Thus, increasing $r$ leads to a higher expressive power in the layers where this is required. 
We analyze both SVDP and STTP on SNGAN: reparameterizing only the discriminator and reparameterizing both the discriminator and the generator.

\paragraph{Reparameterizing Discriminator}

We keep the generator intact and unconstrained; for the discriminator, we apply both SVDP and STTP with $r \in \{32,64\}$. 

Note that the reported performance reflects the quality of the same unconstrained generator, trained together with varying discriminator constraints. 
This can be seen as another form of limiting the discriminator's capacity (by rank instead of the number of features), similar to the reduced discriminator setting (Table~\ref{tbl:results:gan:reduced}).

Table~\ref{tbl:ranks}~(left) shows the results of this group of experiments: $r=64$ gives the best performance, suggesting that it is the optimal low-rank regime for SNGAN-32 in the given settings. 
STTP with $r=64$ has a similar performance to SVDP with $r=32$, with only ${\sim}66\%$ of the number of parameters.

\paragraph{Reparameterizing Generator}

It remains unclear what kind of effect the rank reduction may have on the generator. 
To this end, we choose a sufficiently good discriminator setting with $r=64$ SVDP and explore the same range of ranks with SVDP and STTP, only now varying just the generator's parameterization. 
To allow for the large magnitude of inputs to the last $\mathrm{tanh}$ layer, we leave the first fully-connected layer in its original, unparameterized form.

\begin{figure}[t]
\centering
\includegraphics[trim=0 0 50 70, clip, width=\linewidth]{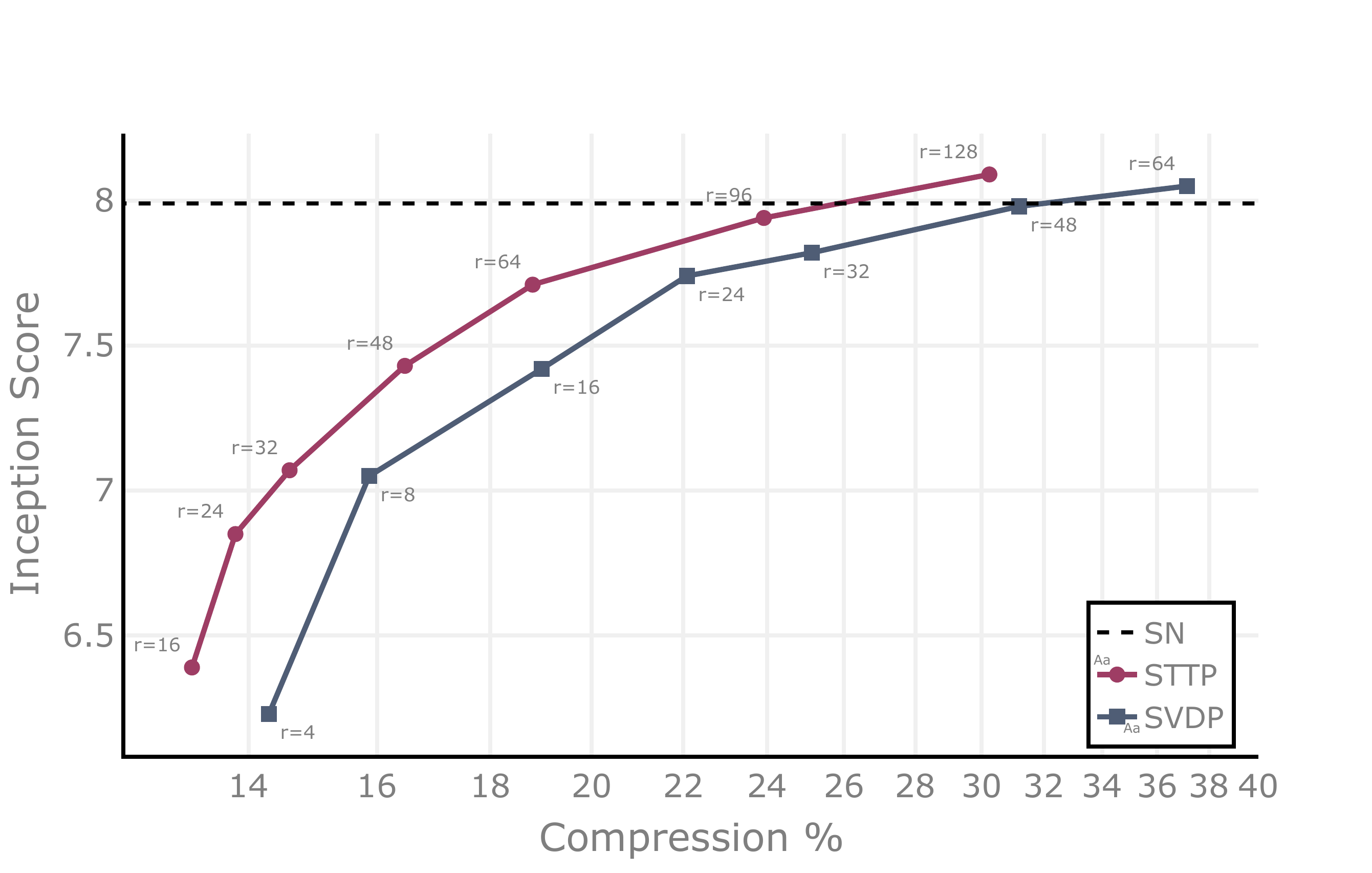}
\caption[]{
Performance-compression curves of the SNGAN generator, while the discriminator is statically parameterized with SVDP $r=64$. Performance is the Inception Score of $\mathcal{G}$ (higher is better); compression is the ratio~\eqref{eq:compression} of the parameterized and the original $\mathcal{G}$ parameter counts (lower is better). Each data point is annotated with the rank $r$ of $\mathcal{G}$ used to produce the score. STTP consistently outperforms SVDP, achieves higher scores with larger ranks and fewer parameters, suggesting a better rank-efficiency of STTP.
}
\label{fig:plot:gan:ranks}
\end{figure}

The results can be seen in Table~\ref{tbl:ranks} (right): STTP is more sensitive to rank reduction than SVDP; however, matching SVDP performance is possible at a lower parameter count with twice a larger rank. 
We additionally visualize the performance-compression frontier (Fig.~\ref{fig:plot:gan:ranks}), which confirms that STTP is more rank-efficient than SVDP in the examined setting and low-rank regime. 

\subsection{Image Classification}

We additionally study the effect of applying both SVDP and STTP in a more traditional image classification setting. 
To this end, we train an image classification CNN on the CIFAR-10 dataset and report the Top-1 accuracy over the validation split. 
For the CNN, we use a Wide ResNet~\citep{wideresnet} with 28 layers and widening factor 10 (WRN-28-10). 
We train for 100K steps with SGD, the initial learning rate 0.1, decaying linearly to 0, momentum 0.9, weight decay 1e-4. 
The spectrum is learned and regularized in all experiments (the '-R' flavor of models).
The first convolutional layer is kept unparameterized to allow for a large global Lipschitz constant required to approximate one-hot distributions in the softmax layer.

\begin{figure}[ht]
\centering
\includegraphics[trim=0 0 68 70, clip, width=\linewidth]{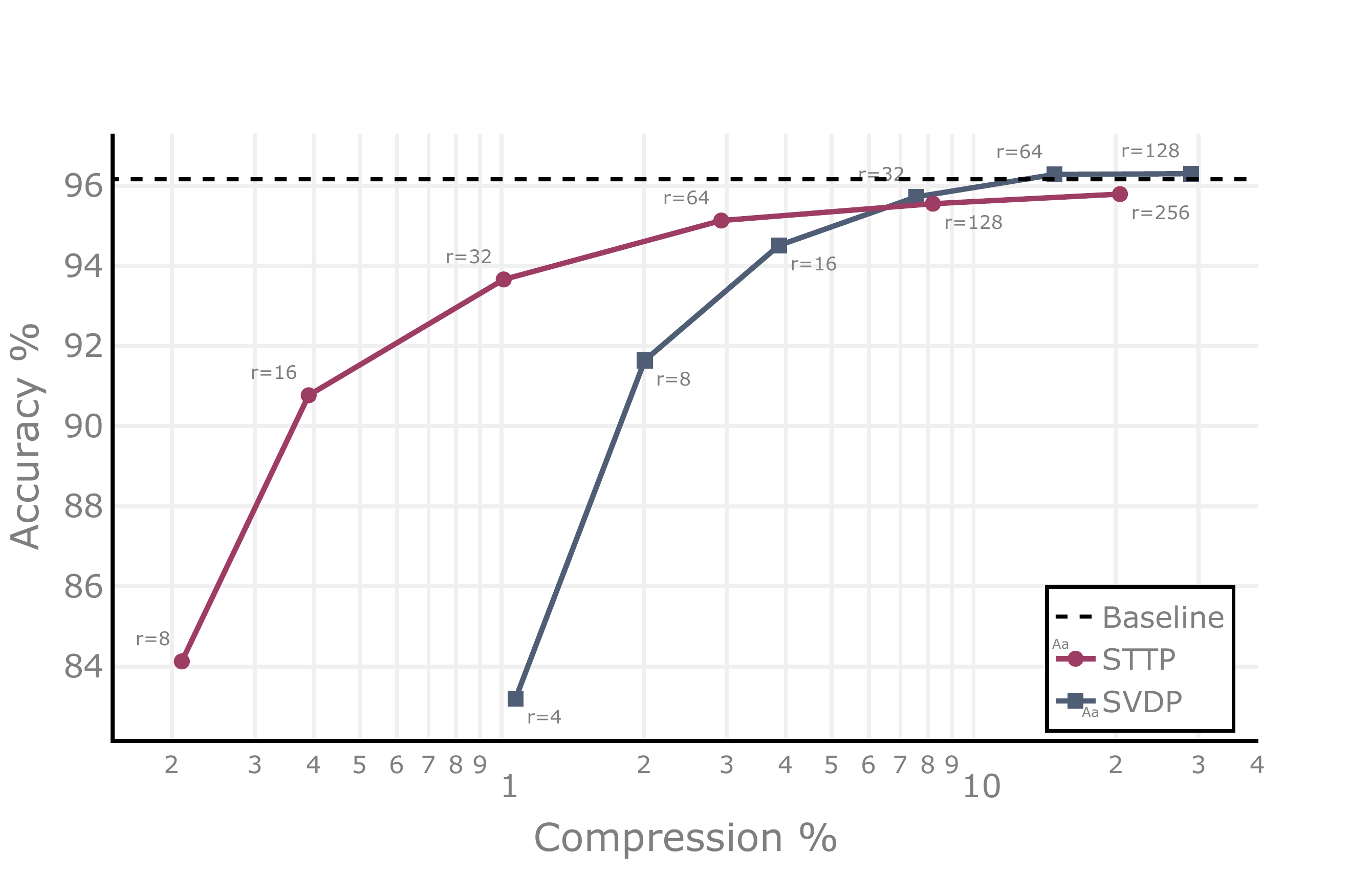}
\caption[]{
Performance-compression curves of the proposed methods in image classification with WRN-28-10 on CIFAR10. 
Performance is the Top-1 Accuracy of the image class prediction model (higher is better); compression is the ratio~\eqref{eq:compression} of the parameterized and the original model's parameter counts (lower is better). 
Each data point is annotated with the respective rank $r$ hyperparameter used within model layers to produce the score. 
STTP outperforms SVDP in the low-rank regime while achieving higher scores with larger ranks.
}
\label{fig:imgcls:wrn}
\end{figure}

As shown in the performance-compression plot in Fig.~\ref{fig:imgcls:wrn}, STTP produces highly compressed models with less than 10\% parameters of the original model, which still achieve competitive performance. 
In this extreme compression regime, STTP clearly outperforms SVDP by a large margin (e.g., 10\% performance gap at a 1\% compression ratio).
However, SVDP is suitable for moderate compression of models, capable of achieving full uncompressed performance.
These observations agree with those made in the GAN setting (Sec.~\ref{lbl:svdvstt}).

\section{Conclusion}
\label{lbl:conclusion}

We presented the Spectral Tensor Train Parameterization (STTP), a novel low-rank parameterization of weight matrices of convolutional and linear layers with embedded spectral properties. 
We analyzed the parameter efficiency of the proposed parameterization in the image classification setting, compared it to the SVD parameterization, and concluded that it permits efficient rank utilization with fewer learned parameters.
Finally, we analyzed our parameterization in the GAN setting and showed that it leads to better training stability. 
Future research directions may include finding optimal per-layer rank selection policies (e.g., using Neural Architecture Search) and analyzing the induced sparsity effect on model biases and rare class performance.

\onecolumn

\section*{Acknowledgements}
This work is funded by Toyota Motor Europe via the research project TRACE-Zurich. 
We thank NVIDIA for GPUs, Amazon Activate for EC2 credits, and Leonhard cluster at ETH Zurich for the compute. 
We also thank Martin Danelljan, Roman Andreev, and anonymous reviewers for the valuable feedback and time spent.
%
%
%
%
\\
\\
\hsize\textwidth\linewidth
\hsize\toptitlebar 
{
  \centering
  {
    \Large
    \bfseries 
    Spectral Tensor Train Parameterization of Deep Learning Layers\\
    Supplementary Materials 
    \par
  }
}
\bottomtitlebar 

\begin{figure}[ht]
\centering
\subfloat[]{%
\label{fig:svd_contraction}{
\begin{tikzpicture}

\node (p4) at (-\httsize,0) {};
\node (p5) at (0,0) {};
\node (p6) at (\httsize,0) {};

\node (phantom) at (0,-4.25*\vsvdsize) {};

\path[-] ($(p4) + (\rad,0)$) edge node[anchor=center, below] {{\footnotesize $r$}} ($(p5) - (\rad,0)$);
\path[-] ($(p5) + (\rad,0)$) edge node[anchor=center, below] {{\footnotesize $r$}} ($(p6) - (\rad,0)$);

\draw[fill] ($(p4) + (-45:\rad)$) arc (-45:135:\rad) -- cycle ;
\draw[fill] (p5)-- ($(p5) + (0:\rad)$) arc (0:360:\rad) -- cycle ;
\draw[fill] ($(p6) + (45:\rad)$) arc (45:225:\rad) -- cycle ;

\draw[thick] (p4) circle (\rad);
\draw[thick] (p5) circle (\rad);
\draw[thick] (p6) circle (\rad);

\path[-] ($(p4) - (0,\rad)$) edge ($(p4) - (0,\rad+\vttsize)$);
\path[-] ($(p6) - (0,\rad)$) edge ($(p6) - (0,\rad+2*\vttsize)$);

\node at ($(p4) - (0,\rad+\vttsize+0.3)$) {\ \ {\small $d_{\dout}$}};
\node at ($(p6) - (-0.3,\rad+\vttsize+0.3)$) {\ \ {\small $d_{\din}$}};

\node at ($(p4) + (0,3.5*\rad)$) {\ $U$};
\node at ($(p5) + (0,3.5*\rad)$) {$\Sigma$};
\node at ($(p6) + (0,3.5*\rad)$) {\ $V^{\top}$};

\draw[thick] ($(p6) - (0,2*\rad+2*\vttsize)$) circle (\rad);
\node at ($(p6) - (0,2*\rad+2*\vttsize) - (0,3*\rad)$) {$x$};

\path[-] ($(p6) - (\rad,2*\rad+2*\vttsize)$) edge ($(p6) - (8*\rad,2*\rad+2*\vttsize)$);
\node at ($(p5) - (1.5*\rad,2*\rad+2*\vttsize)$) {$d_x$};

\end{tikzpicture}
}}\qquad
\subfloat[]{%
\label{fig:tt_contraction}{
\begin{tikzpicture}

\foreach \i in {1,...,9}
{
    \node (p\i) at (\i*\httsize,0) {};
}

\path[-] ($(p1) + (\rad,0)$) edge node[anchor=center, below] {{\footnotesize $R_1^{\dout}$}} ($(p2) - (\rad,0)$);
\path[-] ($(p2) + (\rad,0)$) edge node[anchor=center, below] {{\footnotesize $R_2^{\dout}$}} ($(p3) - (2*\rad,0)$);
\path[-] ($(p3) + (2*\rad,0)$) edge node[anchor=center, below] {{\footnotesize $R_{D_\dout{\texttt{-}}1}^{\dout}$}} ($(p4) - (\rad,0)$);
\path[-] ($(p4) + (\rad,0)$) edge node[anchor=center, below] {{\footnotesize $r$}} ($(p5) - (\rad,0)$);
\path[-] ($(p5) + (\rad,0)$) edge node[anchor=center, below] {{\footnotesize $r$}} ($(p6) - (\rad,0)$);
\path[-] ($(p6) + (\rad,0)$) edge node[anchor=center, below] {{\footnotesize $R_{D_\din{\texttt{-}}1}^{\din}$}} ($(p7) - (2*\rad,0)$);
\path[-] ($(p7) + (2*\rad,0)$) edge node[anchor=center, below] {{\footnotesize $R_2^{\din}$}} ($(p8) - (\rad,0)$);
\path[-] ($(p8) + (\rad,0)$) edge node[anchor=center, below] {{\footnotesize $R_1^{\din}$}} ($(p9) - (\rad,0)$);

\draw[fill] ($(p1) + (-45:\rad)$) arc (-45:135:\rad) -- cycle ;
\draw[fill] ($(p2) + (-45:\rad)$) arc (-45:135:\rad) -- cycle ;
\draw[fill] ($(p4) + (-45:\rad)$) arc (-45:135:\rad) -- cycle ;
\draw[fill] (p5)-- ($(p5) + (0:\rad)$) arc (0:360:\rad) -- cycle ;
\draw[fill] ($(p6) + (45:\rad)$) arc (45:225:\rad) -- cycle ;
\draw[fill] ($(p8) + (45:\rad)$) arc (45:225:\rad) -- cycle ;
\draw[fill] ($(p9) + (45:\rad)$) arc (45:225:\rad) -- cycle ;

\draw[thick] (p1) circle (\rad);
\draw[thick] (p2) circle (\rad);
\node at (p3) {\,\smaller $\cdots$};
\draw[thick] (p4) circle (\rad);
\draw[thick] (p5) circle (\rad);
\draw[thick] (p6) circle (\rad);
\node at (p7) {\,\smaller $\cdots$};
\draw[thick] (p8) circle (\rad);
\draw[thick] (p9) circle (\rad);

\path[-] ($(p1) - (0,\rad)$) edge ($(p1) - (0,\rad+\vttsize)$);
\path[-] ($(p2) - (0,\rad)$) edge ($(p2) - (0,\rad+\vttsize)$);
\path[-] ($(p4) - (0,\rad)$) edge ($(p4) - (0,\rad+\vttsize)$);
\path[-] ($(p6) - (0,\rad)$) edge ($(p6) - (0,\rad+2*\vttsize)$);
\path[-] ($(p8) - (0,\rad)$) edge ($(p8) - (0,\rad+2*\vttsize)$);
\path[-] ($(p9) - (0,\rad)$) edge ($(p9) - (0,\rad+2*\vttsize)$);

\node at ($(p1) - (0,\rad+\vttsize+0.3)$) {\ \ {\small $n^{\dout}_{1}$}};
\node at ($(p2) - (0,\rad+\vttsize+0.3)$) {\ \ {\small $n^{\dout}_{2}$}};
\node at ($(p4) - (0,\rad+\vttsize+0.3)$) {\ \ {\small $n^{\dout}_{D_\dout}$}};
\node at ($(p6) - (-0.4,\rad+\vttsize+0.2)$) {\ \ {\small $n^{\din}_{D_\din}$}};
\node at ($(p8) - (-0.3,\rad+\vttsize+0.2)$) {\ \ {\small $n^{\din}_{2}$}};
\node at ($(p9) - (-0.3,\rad+\vttsize+0.2)$) {\ \ {\small $n^{\din}_{1}$}};

\node at ($(p1) + (0,3.5*\rad)$) {\ \ $\mathcal{U}^{(1)}$};
\node at ($(p2) + (0,3.5*\rad)$) {\ \ $\mathcal{U}^{(2)}$};
\node at ($(p4) + (0,3.5*\rad)$) {\ \ $\mathcal{U}^{(D_\dout)}$};
\node at ($(p5) + (0,3.5*\rad)$) {$\Sigma$};
\node at ($(p6) + (0,3.5*\rad)$) {\ \ $\mathcal{V}^{(D_\din) \top}$};
\node at ($(p8) + (0,3.5*\rad)$) {\ \ $\mathcal{V}^{(2) \top}$};
\node at ($(p9) + (0,3.5*\rad)$) {\ \ $\mathcal{V}^{(1) \top}$};

\draw[rounded corners=8pt]
($(p7) - (2*\rad,\rad+2*\vttsize)$) -- 
($(p6) - (3*\rad,\rad+2*\vttsize)$) -- 
($(p6) - (3*\rad,\rad+3.5*\vttsize)$) -- 
($(p7) - (2*\rad,\rad+3.5*\vttsize)$);
\draw[rounded corners=8pt]
($(p7) - (-2*\rad,\rad+2*\vttsize)$) -- 
($(p9) - (-3*\rad,\rad+2*\vttsize)$) -- 
($(p9) - (-3*\rad,\rad+3.5*\vttsize)$) -- 
($(p7) - (-2*\rad,\rad+3.5*\vttsize)$);
\node at ($(p7) - (0,\rad+2*\vttsize)$) {\,\smaller $\cdots$};
\node at ($(p7) - (0,\rad+3.5*\vttsize)$) {\,\smaller $\cdots$};
\node at ($(p8) - (0.5*\httsize,\rad+2.7*\vttsize)$) {$\widetilde{x}$};
\path[-] ($(p6) - (3*\rad,\rad+2.75*\vttsize)$) edge ($(p6) - (8*\rad,\rad+2.75*\vttsize)$);
\node at ($(p5) - (1.5*\rad,\rad+2.75*\vttsize)$) {$d_x$};
\node[] () [below = 0.5em] at 
($(p8) - (-3*\rad,\rad+3.5*\vttsize)$) 
{};

\end{tikzpicture}
}}
\caption[]{
    Tensor diagrams of a product of (a) SVDP and (b) STTP of a weight matrix $W\in\mathbb{R}^{d_{\dout}\times d_{\din}}$ and an input $x \in \mathbb{R}^{d_\din \times d_x}$, where $d_x$ is a batch or any other second dimension. 
    See legend in Fig.~\ref{fig:networks} caption.
    $\widetilde x$ is obtained by tensorizing $x$ along the dimension $d_\din$ using the same dimension factorization as the corresponding dimension of the matrix $W$. 
    The factorized dimensions of $\widetilde{x}$ are connected with the recipient dimensions of TT-cores $\mathcal{V}^{(k)}$. 
    The contraction order of all connected edges defines FLOPs and memory requirements for computing $y=W x$. 
    After the contraction, dimensions of the output $\widetilde{y}$ can be flattened to recover $y$.
}
\label{fig:contractions}
\end{figure}
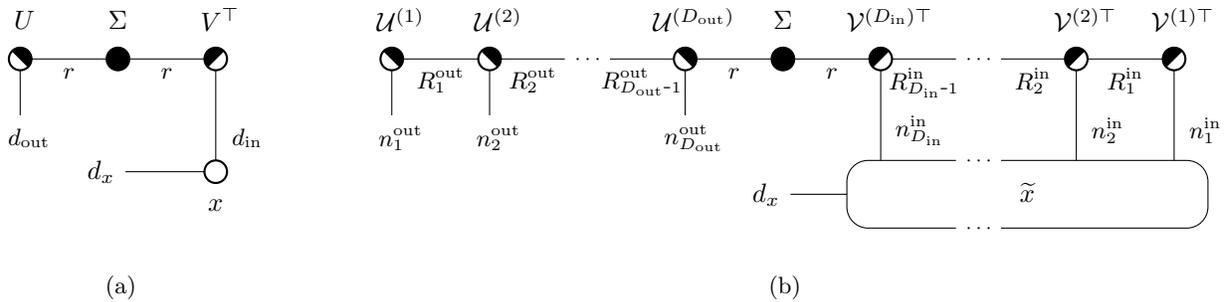

\section{Computational Shortcuts of Low-Rank Affine Mappings}
\label{sec:compshortcuts}

Both SVDP and STTP support ``decompression'' of the weight matrix $W \in \mathbb{R}^{d_{\dout} \times d_{\din}}$, which can be used for computing the mapping $\omega(x)$ directly as $y=Wx$ for some input $x \in \mathbb{R}^{d_{\din} \times d_x}$. 
The last dimension of the input $x$ may correspond to the batch dimension, so its value may be large during neural network training or equal to 1 during inference.
The low-rank structure of the proposed parameterizations allows for taking certain computational shortcuts for computing either $W$ or the mapping output $y$, as measured in floating-point operations (\textit{FLOPs}).

\paragraph{SVDP} 
The number of FLOPs required to decompress $W$ given $U$, $\Sigma$, and $V$ is $r \min(d_\dout, d_\din) + 2 r d_\dout d_\din$. 
Computing $y=Wx$ then takes another $2 d_{\dout} d_{\din} d_x$ FLOPs. 
When $r \ll \min(d_\din,d_\dout)$, and $d_x=1$, computing $y=U (\Sigma (V^\top x))$ following the order indicated by parentheses is preferred. 
Indeed, such computation brings the number of FLOPs down to $r(2d_{\din}+2d_{\dout}+1)$.
Overall, the \textit{optimal contraction order} of a tensor diagram shown in Fig.~\ref{fig:svd_contraction} (which corresponds to arranging parentheses in the expression $U \Sigma V^\top x$) is defined by the sizes of all operands involved in the expression and can be precomputed upon the layer initialization.

\paragraph{STTP}
After computing the TT-cores of matrices $U$ and $V$ from the underlying parameterizations, there are more than two ways to compute the mapping $\omega(x)$.
As before, one can contract the tensor diagram of the matrix $W$ first and then perform the regular computation of $y=Wx$. 
A slightly more efficient way is to contract matrices $U$ and $V$ and then re-use the approach to the low-rank mapping of SVDP. 

Finally, the most efficient approach consists of the following steps: (1) factorization of the first dimension of $x$ into $D_{\din}$ factors: $d_\din=n_1\dots n_{D_\din}$, (2) tensorization of $x$ into a tensor $\widetilde x$ according to the dimension factorization, (3) connecting factorized dimensions of $\widetilde{x}$ with the respective dimensions of the TT-cores $\mathcal{V}$, and finally, (4)~contracting the resulting tensor diagram in Fig.~\ref{fig:tt_contraction} according to the optimal contraction order. 
While finding the optimal contraction order of a generic tensor diagram is an NP-hard problem, efficient algorithms exist for certain classes of graphs~\citep{Smith2018opteinsum}. 
This approach gives us the lowest possible FLOPs count of computing $\omega$ directly in the low-rank space, as both simpler approaches belong to the search space of the contraction order. 
As in the case of SVDP, the optimal contraction order depends on the topology of the tensor diagram and node sizes.
Since the layer dimensions are known in advance, the mapping complexity is not increased at runtime.

\section{Batch Householder Transformation}
\label{lbl:batchedhh}

Computations involving the proposed parameterizations are dominated by orthogonal transformations~\eqref{eq:orgqr}.
In this section, we discuss some aspects that make our approach feasible as the size and the number of neural network layers grow.
Despite Householder transformation being more amenable to SIMD implementation than Givens rotations and matrix exponential maps~\citep{shepard2015representation}, prior works avoid using them altogether due to the lack of framework support\footnote{
  Even though an orthogonal transformation implementing~\eqref{eq:orgqr} can be found in modern automatic differentiation packages as LAPACK bindings ({\normalfont\texttt{?ORGQR}}), these functions rarely support batching or differentiation with respect to inputs.
}, complex implementations, and hardness to scale beyond a handful of layers. 
We overcome these limitations by utilizing a joint parameterization of orthonormal frames of the same size~\citep{obukhov2021torchhouseholder}. 
In the context of SVDP, it allows us to generate multiple orthonormal frames of the same size $d \times r$ (potentially belonging to different layers) in a sequence of a total of $r$ batched Householder reflections. 

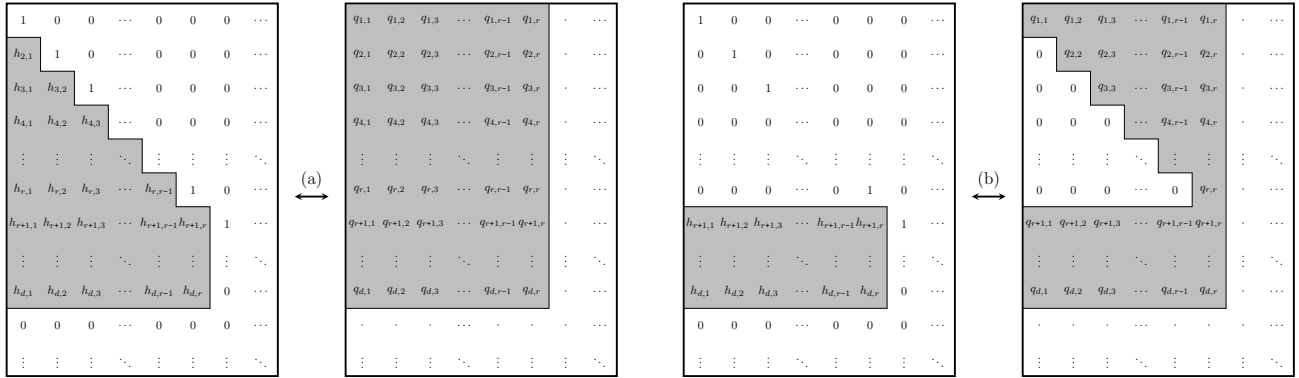
\begin{figure*}[t]
\centering
\resizebox{0.475\linewidth}{!}{%
\begin{tikzpicture}
    \draw [ultra thick, draw=black] 
        (0,9) -- (8,9) -- (8,-2) -- (0,-2) -- cycle;
    \draw [thick, draw=black, fill=gray, fill opacity=0.5]
       (0,8) -- (1,8) -- 
       (1,7) -- (2,7) -- 
       (2,6) -- (3,6) -- 
       (3,5) -- (4,5) -- 
       (4,4) -- (5,4) -- 
       (5,3) -- (6,3) -- 
       (6,0) -- (0,0) -- 
       cycle;

    \node[] at (0.5, 8.5) {$1$};
    \node[] at (0.5, 7.5) {$h_{2,1}$};
    \node[] at (0.5, 6.5) {$h_{3,1}$}; 
    \node[] at (0.5, 5.5) {$h_{4,1}$}; 
    \node[] at (0.5, 4.5) {$\vdots$}; 
    \node[] at (0.5, 3.5) {$h_{r, 1}$}; 
    \node[] at (0.5, 2.5) {$h_{r \Plus 1, 1}$}; 
    \node[] at (0.5, 1.5) {$\vdots$}; 
    \node[] at (0.5, 0.5) {$h_{d, 1}$}; 
    \node[] at (0.5,-0.5) {$0$}; 
    \node[] at (0.5,-1.5) {$\vdots$}; 

    \node[] at (1.5, 8.5) {$0$};
    \node[] at (1.5, 7.5) {$1$};
    \node[] at (1.5, 6.5) {$h_{3,2}$};
    \node[] at (1.5, 5.5) {$h_{4,2}$};
    \node[] at (1.5, 4.5) {$\vdots$}; 
    \node[] at (1.5, 3.5) {$h_{r, 2}$}; 
    \node[] at (1.5, 2.5) {$h_{r \Plus 1, 2}$}; 
    \node[] at (1.5, 1.5) {$\vdots$}; 
    \node[] at (1.5, 0.5) {$h_{d, 2}$}; 
    \node[] at (1.5,-0.5) {$0$}; 
    \node[] at (1.5,-1.5) {$\vdots$}; 

    \node[] at (2.5, 8.5) {$0$};
    \node[] at (2.5, 7.5) {$0$};
    \node[] at (2.5, 6.5) {$1$};
    \node[] at (2.5, 5.5) {$h_{4,3}$}; 
    \node[] at (2.5, 4.5) {$\vdots$}; 
    \node[] at (2.5, 3.5) {$h_{r, 3}$}; 
    \node[] at (2.5, 2.5) {$h_{r \Plus 1, 3}$}; 
    \node[] at (2.5, 1.5) {$\vdots$}; 
    \node[] at (2.5, 0.5) {$h_{d, 3}$}; 
    \node[] at (2.5,-0.5) {$0$}; 
    \node[] at (2.5,-1.5) {$\vdots$}; 

    \node[] at (3.5, 8.5) {$\hdots$};
    \node[] at (3.5, 7.5) {$\hdots$};
    \node[] at (3.5, 6.5) {$\hdots$};
    \node[] at (3.5, 5.5) {$\hdots$};
    \node[] at (3.5, 4.5) {$\ddots$}; 
    \node[] at (3.5, 3.5) {$\hdots$}; 
    \node[] at (3.5, 2.5) {$\hdots$}; 
    \node[] at (3.5, 1.5) {$\ddots$}; 
    \node[] at (3.5, 0.5) {$\hdots$}; 
    \node[] at (3.5,-0.5) {$\hdots$}; 
    \node[] at (3.5,-1.5) {$\ddots$}; 

    \node[] at (4.5, 8.5) {$0$};
    \node[] at (4.5, 7.5) {$0$};
    \node[] at (4.5, 6.5) {$0$};
    \node[] at (4.5, 5.5) {$0$};
    \node[] at (4.5, 4.5) {$\vdots$}; 
    \node[] at (4.5, 3.5) {$h_{r, r \Minus 1}$}; 
    \node[] at (4.5, 2.5) {$h_{r \Plus 1, r \Minus 1}$}; 
    \node[] at (4.5, 1.5) {$\vdots$}; 
    \node[] at (4.5, 0.5) {$h_{d, r \Minus 1}$}; 
    \node[] at (4.5,-0.5) {$0$}; 
    \node[] at (4.5,-1.5) {$\vdots$}; 

    \node[] at (5.5, 8.5) {$0$};
    \node[] at (5.5, 7.5) {$0$};
    \node[] at (5.5, 6.5) {$0$};
    \node[] at (5.5, 5.5) {$0$};
    \node[] at (5.5, 4.5) {$\vdots$}; 
    \node[] at (5.5, 3.5) {$1$}; 
    \node[] at (5.5, 2.5) {$h_{r \Plus 1, r}$}; 
    \node[] at (5.5, 1.5) {$\vdots$}; 
    \node[] at (5.5, 0.5) {$h_{d, r}$}; 
    \node[] at (5.5,-0.5) {$0$}; 
    \node[] at (5.5,-1.5) {$\vdots$}; 

    \node[] at (6.5, 8.5) {$0$};
    \node[] at (6.5, 7.5) {$0$};
    \node[] at (6.5, 6.5) {$0$};
    \node[] at (6.5, 5.5) {$0$};
    \node[] at (6.5, 4.5) {$\vdots$}; 
    \node[] at (6.5, 3.5) {$0$}; 
    \node[] at (6.5, 2.5) {$1$}; 
    \node[] at (6.5, 1.5) {$\vdots$}; 
    \node[] at (6.5, 0.5) {$0$}; 
    \node[] at (6.5,-0.5) {$0$}; 
    \node[] at (6.5,-1.5) {$\vdots$}; 

    \node[] at (7.5, 8.5) {$\hdots$};
    \node[] at (7.5, 7.5) {$\hdots$};
    \node[] at (7.5, 6.5) {$\hdots$};
    \node[] at (7.5, 5.5) {$\hdots$};
    \node[] at (7.5, 4.5) {$\ddots$}; 
    \node[] at (7.5, 3.5) {$\hdots$}; 
    \node[] at (7.5, 2.5) {$\hdots$}; 
    \node[] at (7.5, 1.5) {$\ddots$}; 
    \node[] at (7.5, 0.5) {$\hdots$}; 
    \node[] at (7.5,-0.5) {$\hdots$}; 
    \node[] at (7.5,-1.5) {$\ddots$}; 

    \draw [>=stealth, <->, ultra thick, draw=black] (8.5,3.3) -- (9.5,3.3);
    \node[] at (9,3.8) {\Large(a)};

    \pgfmathtruncatemacro{\offs}{10}

    \draw [ultra thick, draw=black] 
        (\offs+0,9) -- (\offs+8,9) -- (\offs+8,-2) -- (\offs+0,-2) -- cycle;
    \draw [thick, draw=black, fill=gray, fill opacity=0.5]
        (\offs+0,9) -- (\offs+6,9) -- 
        (\offs+6,0) -- (\offs+0,0) -- cycle;
       cycle;

    \node[] at (\offs+0.5, 8.5) {$q_{1,1}$};
    \node[] at (\offs+0.5, 7.5) {$q_{2,1}$};
    \node[] at (\offs+0.5, 6.5) {$q_{3,1}$}; 
    \node[] at (\offs+0.5, 5.5) {$q_{4,1}$}; 
    \node[] at (\offs+0.5, 4.5) {$\vdots$}; 
    \node[] at (\offs+0.5, 3.5) {$q_{r,1}$}; 
    \node[] at (\offs+0.5, 2.5) {$q_{r \Plus 1, 1}$}; 
    \node[] at (\offs+0.5, 1.5) {$\vdots$}; 
    \node[] at (\offs+0.5, 0.5) {$q_{d, 1}$}; 
    \node[] at (\offs+0.5,-0.5) {$\cdot$}; 
    \node[] at (\offs+0.5,-1.5) {$\vdots$}; 

    \node[] at (\offs+1.5, 8.5) {$q_{1,2}$};
    \node[] at (\offs+1.5, 7.5) {$q_{2,2}$};
    \node[] at (\offs+1.5, 6.5) {$q_{3,2}$};
    \node[] at (\offs+1.5, 5.5) {$q_{4,2}$};
    \node[] at (\offs+1.5, 4.5) {$\vdots$}; 
    \node[] at (\offs+1.5, 3.5) {$q_{r,2}$}; 
    \node[] at (\offs+1.5, 2.5) {$q_{r \Plus 1, 2}$}; 
    \node[] at (\offs+1.5, 1.5) {$\vdots$}; 
    \node[] at (\offs+1.5, 0.5) {$q_{d, 2}$}; 
    \node[] at (\offs+1.5,-0.5) {$\cdot$}; 
    \node[] at (\offs+1.5,-1.5) {$\vdots$}; 

    \node[] at (\offs+2.5, 8.5) {$q_{1,3}$};
    \node[] at (\offs+2.5, 7.5) {$q_{2,3}$};
    \node[] at (\offs+2.5, 6.5) {$q_{3,3}$};
    \node[] at (\offs+2.5, 5.5) {$q_{4,3}$}; 
    \node[] at (\offs+2.5, 4.5) {$\vdots$}; 
    \node[] at (\offs+2.5, 3.5) {$q_{r,3}$};
    \node[] at (\offs+2.5, 2.5) {$q_{r \Plus 1, 3}$}; 
    \node[] at (\offs+2.5, 1.5) {$\vdots$}; 
    \node[] at (\offs+2.5, 0.5) {$q_{d, 3}$}; 
    \node[] at (\offs+2.5,-0.5) {$\cdot$}; 
    \node[] at (\offs+2.5,-1.5) {$\vdots$}; 

    \node[] at (\offs+3.5, 8.5) {$\hdots$};
    \node[] at (\offs+3.5, 7.5) {$\hdots$};
    \node[] at (\offs+3.5, 6.5) {$\hdots$};
    \node[] at (\offs+3.5, 5.5) {$\hdots$};
    \node[] at (\offs+3.5, 4.5) {$\ddots$}; 
    \node[] at (\offs+3.5, 3.5) {$\hdots$}; 
    \node[] at (\offs+3.5, 2.5) {$\hdots$}; 
    \node[] at (\offs+3.5, 1.5) {$\ddots$}; 
    \node[] at (\offs+3.5, 0.5) {$\hdots$}; 
    \node[] at (\offs+3.5,-0.5) {$\hdots$}; 
    \node[] at (\offs+3.5,-1.5) {$\ddots$}; 

    \node[] at (\offs+4.5, 8.5) {$q_{1,r \Minus 1}$};
    \node[] at (\offs+4.5, 7.5) {$q_{2,r \Minus 1}$};
    \node[] at (\offs+4.5, 6.5) {$q_{3,r \Minus 1}$};
    \node[] at (\offs+4.5, 5.5) {$q_{4,r \Minus 1}$};
    \node[] at (\offs+4.5, 4.5) {$\vdots$}; 
    \node[] at (\offs+4.5, 3.5) {$q_{r,r \Minus 1}$}; 
    \node[] at (\offs+4.5, 2.5) {$q_{r \Plus 1, r \Minus 1}$}; 
    \node[] at (\offs+4.5, 1.5) {$\vdots$}; 
    \node[] at (\offs+4.5, 0.5) {$q_{d, r \Minus 1}$}; 
    \node[] at (\offs+4.5,-0.5) {$\cdot$}; 
    \node[] at (\offs+4.5,-1.5) {$\vdots$}; 

    \node[] at (\offs+5.5, 8.5) {$q_{1,r}$};
    \node[] at (\offs+5.5, 7.5) {$q_{2,r}$};
    \node[] at (\offs+5.5, 6.5) {$q_{3,r}$};
    \node[] at (\offs+5.5, 5.5) {$q_{4,r}$};
    \node[] at (\offs+5.5, 4.5) {$\vdots$}; 
    \node[] at (\offs+5.5, 3.5) {$q_{r,r}$};
    \node[] at (\offs+5.5, 2.5) {$q_{r \Plus 1, r}$}; 
    \node[] at (\offs+5.5, 1.5) {$\vdots$}; 
    \node[] at (\offs+5.5, 0.5) {$q_{d, r}$}; 
    \node[] at (\offs+5.5,-0.5) {$\cdot$}; 
    \node[] at (\offs+5.5,-1.5) {$\vdots$}; 

    \node[] at (\offs+6.5, 8.5) {$\cdot$};
    \node[] at (\offs+6.5, 7.5) {$\cdot$};
    \node[] at (\offs+6.5, 6.5) {$\cdot$};
    \node[] at (\offs+6.5, 5.5) {$\cdot$};
    \node[] at (\offs+6.5, 4.5) {$\vdots$}; 
    \node[] at (\offs+6.5, 3.5) {$\cdot$}; 
    \node[] at (\offs+6.5, 2.5) {$\cdot$}; 
    \node[] at (\offs+6.5, 1.5) {$\vdots$}; 
    \node[] at (\offs+6.5, 0.5) {$\cdot$}; 
    \node[] at (\offs+6.5,-0.5) {$\cdot$}; 
    \node[] at (\offs+6.5,-1.5) {$\vdots$}; 

    \node[] at (\offs+7.5, 8.5) {$\hdots$};
    \node[] at (\offs+7.5, 7.5) {$\hdots$};
    \node[] at (\offs+7.5, 6.5) {$\hdots$};
    \node[] at (\offs+7.5, 5.5) {$\hdots$};
    \node[] at (\offs+7.5, 4.5) {$\ddots$}; 
    \node[] at (\offs+7.5, 3.5) {$\hdots$}; 
    \node[] at (\offs+7.5, 2.5) {$\hdots$}; 
    \node[] at (\offs+7.5, 1.5) {$\ddots$}; 
    \node[] at (\offs+7.5, 0.5) {$\hdots$}; 
    \node[] at (\offs+7.5,-0.5) {$\hdots$}; 
    \node[] at (\offs+7.5,-1.5) {$\ddots$}; 
    
\end{tikzpicture}%
%
}
\hfill
\resizebox{0.475\linewidth}{!}{%
\begin{tikzpicture}
    \draw [ultra thick, draw=black] 
        (0,9) -- (8,9) -- (8,-2) -- (0,-2) -- cycle;
    \draw [thick, draw=black, fill=gray, fill opacity=0.5]
       (0,3) -- (6,3) -- 
       (6,0) -- (0,0) -- 
       cycle;

    \node[] at (0.5, 8.5) {$1$};
    \node[] at (0.5, 7.5) {$0$};
    \node[] at (0.5, 6.5) {$0$}; 
    \node[] at (0.5, 5.5) {$0$}; 
    \node[] at (0.5, 4.5) {$\vdots$}; 
    \node[] at (0.5, 3.5) {$0$}; 
    \node[] at (0.5, 2.5) {$h_{r \Plus 1, 1}$}; 
    \node[] at (0.5, 1.5) {$\vdots$}; 
    \node[] at (0.5, 0.5) {$h_{d, 1}$}; 
    \node[] at (0.5,-0.5) {$0$}; 
    \node[] at (0.5,-1.5) {$\vdots$}; 

    \node[] at (1.5, 8.5) {$0$};
    \node[] at (1.5, 7.5) {$1$};
    \node[] at (1.5, 6.5) {$0$};
    \node[] at (1.5, 5.5) {$0$};
    \node[] at (1.5, 4.5) {$\vdots$}; 
    \node[] at (1.5, 3.5) {$0$}; 
    \node[] at (1.5, 2.5) {$h_{r \Plus 1, 2}$}; 
    \node[] at (1.5, 1.5) {$\vdots$}; 
    \node[] at (1.5, 0.5) {$h_{d, 2}$}; 
    \node[] at (1.5,-0.5) {$0$}; 
    \node[] at (1.5,-1.5) {$\vdots$}; 

    \node[] at (2.5, 8.5) {$0$};
    \node[] at (2.5, 7.5) {$0$};
    \node[] at (2.5, 6.5) {$1$};
    \node[] at (2.5, 5.5) {$0$}; 
    \node[] at (2.5, 4.5) {$\vdots$}; 
    \node[] at (2.5, 3.5) {$0$}; 
    \node[] at (2.5, 2.5) {$h_{r \Plus 1, 3}$}; 
    \node[] at (2.5, 1.5) {$\vdots$}; 
    \node[] at (2.5, 0.5) {$h_{d, 3}$}; 
    \node[] at (2.5,-0.5) {$0$}; 
    \node[] at (2.5,-1.5) {$\vdots$}; 

    \node[] at (3.5, 8.5) {$\hdots$};
    \node[] at (3.5, 7.5) {$\hdots$};
    \node[] at (3.5, 6.5) {$\hdots$};
    \node[] at (3.5, 5.5) {$\hdots$};
    \node[] at (3.5, 4.5) {$\ddots$}; 
    \node[] at (3.5, 3.5) {$\hdots$}; 
    \node[] at (3.5, 2.5) {$\hdots$}; 
    \node[] at (3.5, 1.5) {$\ddots$}; 
    \node[] at (3.5, 0.5) {$\hdots$}; 
    \node[] at (3.5,-0.5) {$\hdots$}; 
    \node[] at (3.5,-1.5) {$\ddots$}; 

    \node[] at (4.5, 8.5) {$0$};
    \node[] at (4.5, 7.5) {$0$};
    \node[] at (4.5, 6.5) {$0$};
    \node[] at (4.5, 5.5) {$0$};
    \node[] at (4.5, 4.5) {$\vdots$}; 
    \node[] at (4.5, 3.5) {$0$}; 
    \node[] at (4.5, 2.5) {$h_{r \Plus 1, r \Minus 1}$}; 
    \node[] at (4.5, 1.5) {$\vdots$}; 
    \node[] at (4.5, 0.5) {$h_{d, r \Minus 1}$}; 
    \node[] at (4.5,-0.5) {$0$}; 
    \node[] at (4.5,-1.5) {$\vdots$}; 

    \node[] at (5.5, 8.5) {$0$};
    \node[] at (5.5, 7.5) {$0$};
    \node[] at (5.5, 6.5) {$0$};
    \node[] at (5.5, 5.5) {$0$};
    \node[] at (5.5, 4.5) {$\vdots$}; 
    \node[] at (5.5, 3.5) {$1$}; 
    \node[] at (5.5, 2.5) {$h_{r \Plus 1, r}$}; 
    \node[] at (5.5, 1.5) {$\vdots$}; 
    \node[] at (5.5, 0.5) {$h_{d, r}$}; 
    \node[] at (5.5,-0.5) {$0$}; 
    \node[] at (5.5,-1.5) {$\vdots$}; 

    \node[] at (6.5, 8.5) {$0$};
    \node[] at (6.5, 7.5) {$0$};
    \node[] at (6.5, 6.5) {$0$};
    \node[] at (6.5, 5.5) {$0$};
    \node[] at (6.5, 4.5) {$\vdots$}; 
    \node[] at (6.5, 3.5) {$0$}; 
    \node[] at (6.5, 2.5) {$1$}; 
    \node[] at (6.5, 1.5) {$\vdots$}; 
    \node[] at (6.5, 0.5) {$0$}; 
    \node[] at (6.5,-0.5) {$0$}; 
    \node[] at (6.5,-1.5) {$\vdots$}; 

    \node[] at (7.5, 8.5) {$\hdots$};
    \node[] at (7.5, 7.5) {$\hdots$};
    \node[] at (7.5, 6.5) {$\hdots$};
    \node[] at (7.5, 5.5) {$\hdots$};
    \node[] at (7.5, 4.5) {$\ddots$}; 
    \node[] at (7.5, 3.5) {$\hdots$}; 
    \node[] at (7.5, 2.5) {$\hdots$}; 
    \node[] at (7.5, 1.5) {$\ddots$}; 
    \node[] at (7.5, 0.5) {$\hdots$}; 
    \node[] at (7.5,-0.5) {$\hdots$}; 
    \node[] at (7.5,-1.5) {$\ddots$}; 

    \draw [>=stealth, <->, ultra thick, draw=black] (8.5,3.3) -- (9.5,3.3);
    \node[] at (9,3.8) {\Large(b)};

    \pgfmathtruncatemacro{\offs}{10}

    \draw [ultra thick, draw=black] 
        (\offs+0,9) -- (\offs+8,9) -- (\offs+8,-2) -- (\offs+0,-2) -- cycle;
    \draw [thick, draw=black, fill=gray, fill opacity=0.5]
       (\offs+0,8) -- (\offs+1,8) -- 
       (\offs+1,7) -- (\offs+2,7) -- 
       (\offs+2,6) -- (\offs+3,6) -- 
       (\offs+3,5) -- (\offs+4,5) -- 
       (\offs+4,4) -- (\offs+5,4) -- 
       (\offs+5,3) -- (\offs+0,3) -- 
       (\offs+0,0) -- (\offs+6,0) -- 
       (\offs+6,9) -- (\offs+0,9) -- 
       (\offs+0,8) --
       cycle;

    \node[] at (\offs+0.5, 8.5) {$q_{1,1}$};
    \node[] at (\offs+0.5, 7.5) {$0$};
    \node[] at (\offs+0.5, 6.5) {$0$}; 
    \node[] at (\offs+0.5, 5.5) {$0$}; 
    \node[] at (\offs+0.5, 4.5) {$\vdots$}; 
    \node[] at (\offs+0.5, 3.5) {$0$}; 
    \node[] at (\offs+0.5, 2.5) {$q_{r \Plus 1, 1}$}; 
    \node[] at (\offs+0.5, 1.5) {$\vdots$}; 
    \node[] at (\offs+0.5, 0.5) {$q_{d, 1}$}; 
    \node[] at (\offs+0.5,-0.5) {$\cdot$}; 
    \node[] at (\offs+0.5,-1.5) {$\vdots$}; 

    \node[] at (\offs+1.5, 8.5) {$q_{1,2}$};
    \node[] at (\offs+1.5, 7.5) {$q_{2,2}$};
    \node[] at (\offs+1.5, 6.5) {$0$};
    \node[] at (\offs+1.5, 5.5) {$0$};
    \node[] at (\offs+1.5, 4.5) {$\vdots$}; 
    \node[] at (\offs+1.5, 3.5) {$0$}; 
    \node[] at (\offs+1.5, 2.5) {$q_{r \Plus 1, 2}$}; 
    \node[] at (\offs+1.5, 1.5) {$\vdots$}; 
    \node[] at (\offs+1.5, 0.5) {$q_{d, 2}$}; 
    \node[] at (\offs+1.5,-0.5) {$\cdot$}; 
    \node[] at (\offs+1.5,-1.5) {$\vdots$}; 

    \node[] at (\offs+2.5, 8.5) {$q_{1,3}$};
    \node[] at (\offs+2.5, 7.5) {$q_{2,3}$};
    \node[] at (\offs+2.5, 6.5) {$q_{3,3}$};
    \node[] at (\offs+2.5, 5.5) {$0$}; 
    \node[] at (\offs+2.5, 4.5) {$\vdots$}; 
    \node[] at (\offs+2.5, 3.5) {$0$};
    \node[] at (\offs+2.5, 2.5) {$q_{r \Plus 1, 3}$}; 
    \node[] at (\offs+2.5, 1.5) {$\vdots$}; 
    \node[] at (\offs+2.5, 0.5) {$q_{d, 3}$}; 
    \node[] at (\offs+2.5,-0.5) {$\cdot$}; 
    \node[] at (\offs+2.5,-1.5) {$\vdots$}; 

    \node[] at (\offs+3.5, 8.5) {$\hdots$};
    \node[] at (\offs+3.5, 7.5) {$\hdots$};
    \node[] at (\offs+3.5, 6.5) {$\hdots$};
    \node[] at (\offs+3.5, 5.5) {$\hdots$};
    \node[] at (\offs+3.5, 4.5) {$\ddots$}; 
    \node[] at (\offs+3.5, 3.5) {$\hdots$}; 
    \node[] at (\offs+3.5, 2.5) {$\hdots$}; 
    \node[] at (\offs+3.5, 1.5) {$\ddots$}; 
    \node[] at (\offs+3.5, 0.5) {$\hdots$}; 
    \node[] at (\offs+3.5,-0.5) {$\hdots$}; 
    \node[] at (\offs+3.5,-1.5) {$\ddots$}; 

    \node[] at (\offs+4.5, 8.5) {$q_{1,r \Minus 1}$};
    \node[] at (\offs+4.5, 7.5) {$q_{2,r \Minus 1}$};
    \node[] at (\offs+4.5, 6.5) {$q_{3,r \Minus 1}$};
    \node[] at (\offs+4.5, 5.5) {$q_{4,r \Minus 1}$};
    \node[] at (\offs+4.5, 4.5) {$\vdots$}; 
    \node[] at (\offs+4.5, 3.5) {$0$}; 
    \node[] at (\offs+4.5, 2.5) {$q_{r \Plus 1, r \Minus 1}$}; 
    \node[] at (\offs+4.5, 1.5) {$\vdots$}; 
    \node[] at (\offs+4.5, 0.5) {$q_{d, r \Minus 1}$}; 
    \node[] at (\offs+4.5,-0.5) {$\cdot$}; 
    \node[] at (\offs+4.5,-1.5) {$\vdots$}; 

    \node[] at (\offs+5.5, 8.5) {$q_{1,r}$};
    \node[] at (\offs+5.5, 7.5) {$q_{2,r}$};
    \node[] at (\offs+5.5, 6.5) {$q_{3,r}$};
    \node[] at (\offs+5.5, 5.5) {$q_{4,r}$};
    \node[] at (\offs+5.5, 4.5) {$\vdots$}; 
    \node[] at (\offs+5.5, 3.5) {$q_{r,r}$};
    \node[] at (\offs+5.5, 2.5) {$q_{r \Plus 1, r}$}; 
    \node[] at (\offs+5.5, 1.5) {$\vdots$}; 
    \node[] at (\offs+5.5, 0.5) {$q_{d, r}$}; 
    \node[] at (\offs+5.5,-0.5) {$\cdot$}; 
    \node[] at (\offs+5.5,-1.5) {$\vdots$}; 

    \node[] at (\offs+6.5, 8.5) {$\cdot$};
    \node[] at (\offs+6.5, 7.5) {$\cdot$};
    \node[] at (\offs+6.5, 6.5) {$\cdot$};
    \node[] at (\offs+6.5, 5.5) {$\cdot$};
    \node[] at (\offs+6.5, 4.5) {$\vdots$}; 
    \node[] at (\offs+6.5, 3.5) {$\cdot$}; 
    \node[] at (\offs+6.5, 2.5) {$\cdot$}; 
    \node[] at (\offs+6.5, 1.5) {$\vdots$}; 
    \node[] at (\offs+6.5, 0.5) {$\cdot$}; 
    \node[] at (\offs+6.5,-0.5) {$\cdot$}; 
    \node[] at (\offs+6.5,-1.5) {$\vdots$}; 

    \node[] at (\offs+7.5, 8.5) {$\hdots$};
    \node[] at (\offs+7.5, 7.5) {$\hdots$};
    \node[] at (\offs+7.5, 6.5) {$\hdots$};
    \node[] at (\offs+7.5, 5.5) {$\hdots$};
    \node[] at (\offs+7.5, 4.5) {$\ddots$}; 
    \node[] at (\offs+7.5, 3.5) {$\hdots$}; 
    \node[] at (\offs+7.5, 2.5) {$\hdots$}; 
    \node[] at (\offs+7.5, 1.5) {$\ddots$}; 
    \node[] at (\offs+7.5, 0.5) {$\hdots$}; 
    \node[] at (\offs+7.5,-0.5) {$\hdots$}; 
    \node[] at (\offs+7.5,-1.5) {$\ddots$}; 
    
\end{tikzpicture}%
%
}
\caption[]{
Visualization of the Padded Householder parameterization of orthonormal frames: (a) Full parameterization, (b) Reduced parameterization. 
Shaded areas with $h_{i,j}$ values represent parameters of reflectors; $q_{i,j}$ values represent orthonormal frame elements, affected by the parameterization. 
The padded parameterization allows for more efficient batching of orthonormal frames of different sizes for better parallelism at the expense of memory.
}
\label{fig:paddedhh}
\end{figure*}

In the context of STTP, all matricized TT-cores are represented as orthonormal frames of a limited set of sizes and can be computed independently of each other. 
Specifically, the sizes are of the form $R_a n_b \times R_c$, where $n_b$ belongs to the set of all possible factors of weight matrices' sizes in the whole network, and $R_a, R_c$ belong to the set of all possible TT-rank values induced by matrix dimensions and rank $r$. 
In practice, we can reduce the set of different sizes of orthonormal frames used in the model by following high-level design recommendations discussed in Sec.~\ref{lbl:sec:architecturedesign}.
These observations lead to the improvement of TT-cores computation parallelism by having fewer different orthonormal frame sizes and a higher number of frames in each batch of a fixed size.

\paragraph{Padded Householder}
Although batching orthonormal frames of the same size improves parallelism, batches of different sizes are still processed in sequence.
Here we show how to trade memory for parallelism and perform parameterizations of multiple orthonormal frames of different sizes in a single batch. 
Concretely, given a set of sizes $d_1 \times r_1, \dots, d_k \times r_k$, we parameterize the respective orthonormal frames using the proposed Padded Householder parameterization in a batch of $k$ matrices of size $\mathrm{max}(d_1,\dots,d_k) \times \mathrm{max}(r_1,\dots,r_k)$, as shown in Fig.~\ref{fig:paddedhh}\hyperref[fig:paddedhh]{a}. 
Indeed, by zeroing $h_{p,i,j}: \forall i>d_p, \forall j>r_p$ in the $p$-th matrix of parameters, propagating $1$ on the diagonal for $j>r_p$, and applying the Householder transformation, the resulting leading sub-matrix is in $\mathrm{St}(d_p,r_p), p=1\ldots k$.

Householder parameterization padding can also be used together with the Reduced parameterization introduced in Sec.~\ref{sec:param_stiefel_full} (Fig.~\ref{fig:paddedhh}\hyperref[fig:paddedhh]{b}) to describe elements in $\mathrm{St}_\mathsf{U}(d,r)$. 
Indeed, both Reduced and Padded variations employ the same transformation and differ only in the placement of constants $\{0,1\}$. 
Thus it is possible to perform parameterization of all orthonormal frames needed by the model in a single batch of rank-$r$ orthonormal frames.

\section{Neural Architecture Design for Efficient Batching of Orthonormal Frames}
\label{lbl:sec:architecturedesign}

Sec.~\ref{lbl:batchedhh} points out the possibility of parallel computation of orthonormal frames, potentially belonging to different layers (and cores in STTP). 
The ability to compute most of the orthonormal frames of weight matrices in parallel is the defining factor of the compute throughput during training. 
There are a few neural architecture design traits, which have a direct impact on the effectiveness of such batching with or without padding. 

Recall that SVDP of a 2D convolution with the weight matrix $W\in\mathbb{R}^{C_\mathrm{out}\times C_\mathrm{in}K^2}$ requires the computation of orthonormal frames $U\in \mathbb{R}^{C_\mathrm{out}\times R}$ and $V\in \mathbb{R}^{C_\mathrm{in}K^2\times R}$, where $R=\min(r, C_\mathrm{out}, C_\mathrm{in}K^2)$. 
To ensure that orthonormal frames from different layers can be batched, one should aim to reduce the amount of variation in the dimensions of matrices $U$ and $V$ belonging to different layers. 
For example, this is achieved with most residual architectures such as~\cite{resnet,wideresnet}, which contain repetitions of residual blocks. 
Each unique orthonormal frame size forms a separate batch, which in turn requires a separate function call during training. Such cases include, for example, components of the preamble layer attaching to RGB inputs or layers with unique $d_\dout$ or $d_\din$ less than $r$, causing rank demotion to satisfy the constraint $r \leq \min(d_\dout, d_\din)$.

STTP declares less strict constraints on the overall network architecture and weight matrix sizes than SVDP. 
Recall the TT parameterization of an orthonormal frame $U\in \mathbb{R}^{d \times r}$ requires computing dimension factorization $d=n_1\dots n_D$ for some $n_1,\dots, n_D \in \mathbb{N} \setminus \{1\}$, for example, prime factors of $d$ with repetition. 
Then $U$ can have a low-rank parameterization through a number of TT-cores~\eqref{eq:tt} with matricized dimensions $R_{k-1}n_k \times R_k$ for $k=1,\dots,D$. 
The ranks $R_k$ are defined as $\min(r, R_k^{\mathrm{max}})$~\eqref{eq:ttrankcapprelim}. 
To give a concrete example, consider a convolutional layer with the weight matrix $W\in \mathbb{R}^{16\times8\cdot 3 \cdot 3}$, and $r=4$. 
Then matrices $U \in \mathbb{R}^{16\times4}$ and $V^\top \in \mathbb{R}^{8\cdot3\cdot3 \times 4}$ will be tensorized into tensors $\widetilde{U}\in \mathbb{R}^{2\times2\times2\times2\times4}$ and $\widetilde{V}\in \mathbb{R}^{2\times2\times2\times3\times3\times4}$ using the prime factors of the first dimensions of matrices $U$ and $V^\top$. 
The TT-rank and dimensions of $\widetilde W$ induced by such dimensions factorization~\eqref{eq:ttdof} will be $R=(1,2,4,4,4,4,4,4,2,1)$, $n=(2,2,2,2,3,3,2,2,2)$, and the complete set of matricized core sizes will contain: $\{(2\times2)^2, (4\times4)^2, (8\times4)^3, (12\times4)^2\}$ (upper index indicates the size of the batch). 
Thus, the relation of $r$ to the size of the layer and the ability to factorize dimensions of the weight matrices play crucial roles in reducing the variation of orthonormal frame sizes involved in the parameterization. 
With this in mind, here are a few neural architecture design rules for maximum computation throughput and efficiency with STTP:
\begin{itemize}
    \item $r$ should be substantially smaller than the maximum dimension of a weight matrix in the whole network (e.g., $r=64$ with 512 features in the largest layer);
    \item the set of convolutional filter sizes (e.g., $\{1, 3\}$) in the entire network should be small;
    \item usage of large (e.g., greater than 3) prime factors should be avoided in $r$, channel, and filter sizes;
    \item best throughput can be achieved with $r$, channel, and filter sizes being powers of a small factor (e.g., 2 or 3).
\end{itemize}

\section{Training Considerations}

\paragraph{Optimizer weight decay, L2 regularization}
The role of regularization with SVDP and STTP is fundamentally different from the regularization of the regular affine layers.
Whereas the latter results in simpler models due to the reduction of the Frobenius norm of weight matrices, the former will reduce individual reflectors' magnitudes, which promotes a truncated diagonal structure in weight matrices.
This may or may not be the desired effect, depending on higher-level design decisions, such as the presence of skip connections; this topic is well beyond the scope of the current work.
Frobenius regularization of the parameterized weight matrices can still be implemented by simply imposing an L2 penalty or performing weight decay of the learned singular values.

\paragraph{Initialization}
Most weight matrix initialization schemes in deep learning are motivated by norm preservation of the layer mapping~\citep{heinit}.
While SVDP and STTP achieve the same goal through the embedded spectral properties, a good initialization still plays an important role in the convergence speed.
We experimented with three different ways of initializing orthonormal frames in both SVDP and STTP:
(1) truncated identity matrix $I_{d\times r}$;
(2) orthogonal initialization with QR decomposition of a random normal matrix~\citep{saxe2013exact}: $\mathrm{QR}(N_{d\times r})$, where elements of $N_{d\times r}$ are i.i.d. sampled from $\mathcal{N}(0, 1)$;
(3) orthogonal initialization with a noisy identity matrix $\mathrm{QR}(I_{d\times r} + \alpha N_{d\times r})$. 
All initialization schemes resulted in a good model performance at the end of the training; however, the noisy identity scheme with $\alpha=1e{-}4$ exhibited faster convergence in the considered experiments with SNGAN.
We conjecture that the best value of $\alpha$ depends on the dimensions $d$ and $r$.

\section{Proof of Proposition~\ref{prop:ttorth}}
\label{sec:proof1}

The fact that $\mathcal{M} (\mathcal{U}^{(k)})\in \mathrm{St}(R_{k-1}n_k, R_k)$ implies $I_{R_k} = \mathcal{M} (\mathcal{U}^{(k)})^\top \mathcal{M} (\mathcal{U}^{(k)})$, or in the index notation,
\begin{equation}
\label{eq:app:orth}
   \delta_{\mu\nu} = 
   \sum_{\beta_{k-1}=1}^{R_{k-1}} \sum_{i_k=1}^{n_k} 
   \mathcal{M} (\mathcal{U}^{(k)})_{\overline{\beta_{k-1} i_k},\mu}
   \,
   \mathcal{M} (\mathcal{U}^{(k)})_{\overline{\beta_{k-1} i_k},\nu}
   =
   \sum_{\beta_{k-1}=1}^{R_{k-1}} \sum_{i_k=1}^{n_k} 
   \mathcal{U}^{(k)}_{\beta_{k-1},i_k,\mu} \, \mathcal{U}^{(k)}_{\beta_{k-1},i_k,\nu},
\end{equation}
where $\delta_{\mu\nu}$ is the Kronecker delta.
To show that $\mathcal{T}(\mathcal{U}^{(1)},\dots,\mathcal{U}^{(D)})\in\mathrm{St}(n_1 \cdots n_D, r)$, let us write the orthogonality condition in index notation (for the ease of notation, we omit ranges in which indices vary):
\begin{equation}
\label{eq:app:orth2}
\begin{split}
    \sum_{i_1,\dots,i_D}
    \mathcal{T}(\mathcal{U}^{(1)},&\dots,\mathcal{U}^{(D)})_{\overline{i_1\dots,i_D},\mu} \ 
     \mathcal{T}(\mathcal{U}^{(1)},\dots,\mathcal{U}^{(D)})_{\overline{i_1\dots,i_D},\nu}  
     \\
     =\sum_{i_1,\dots,i_D} 
        &\left(
            \sum_{\beta_0,\dots,\beta_{D-1}}
                \mathcal{U}^{(1)}_{\beta_0, i_1,\beta_1}
                \mathcal{U}^{(2)}_{\beta_1,i_2,\beta_2}
                \cdots
                \mathcal{U}^{(D)}_{\beta_{D\text{-}1},i_{D}, \mu}
        \right)
        \left(
            \sum_{\beta_0,\dots,\beta_{D-1}}
                \mathcal{U}^{(1)}_{\beta_0, i_1,\beta_1}
                \mathcal{U}^{(2)}_{\beta_1,i_2,\beta_2}
                \cdots
                \mathcal{U}^{(D)}_{\beta_{D\text{-}1},i_{D}, \nu}
        \right)  \\
        &=\sum_{i_1,\dots,i_D}
        \sum_{\beta_0,\dots,\beta_{D-1}}
        \sum_{\tilde\beta_0,\dots,\tilde\beta_{D-1}}
            \mathcal{U}^{(1)}_{\beta_0, i_1,\beta_1}
            \mathcal{U}^{(1)}_{\tilde\beta_0, i_1,\tilde\beta_1}
            \cdots
            \mathcal{U}^{(D)}_{\beta_{D\text{-}1},i_{D}, \mu}
            \mathcal{U}^{(D)}_{\tilde\beta_{D\text{-}1},i_{D}, \nu},
\end{split}
\end{equation}
and note that $\beta_0$ and $\tilde\beta_0$ vary from $1$ to $1$, so with~\eqref{eq:app:orth} for $k=1$, we get
\[
        \sum_{i_1} \mathcal{U}^{(1)}_{1, i_1,\beta_1} \mathcal{U}^{(1)}_{1, i_1,\tilde\beta_1} = \delta_{\beta_1\tilde\beta_1}.
\]
The latter expression implies that in the last line of~\eqref{eq:app:orth2}, after summing over $i_1$, only the terms with $\tilde\beta_1 = \beta_1$ remain.
We can now apply~\eqref{eq:app:orth} for $k=2$:
\[
    \sum_{\beta_1, i_2}\mathcal{U}^{(2)}_{\beta_{1},i_2,\beta_2} \, \mathcal{U}^{(2)}_{\beta_{1},i_2,\tilde\beta_2} = \delta_{\beta_2 \tilde\beta_2}.
\]
Proceeding recursively, we obtain that~\eqref{eq:app:orth2} equals $\delta_{\mu\nu}$, which completes the proof.

\section{Proof of Proposition~\ref{prop:ttredundancy}}
\label{sec:proof2}

Let us first show that
$
    \mathcal{T}\left(\mathcal{U}^{(1)}, \mathcal{U}^{(2)},\dots,\mathcal{U}^{(D)}\right) =  \mathcal{T}\left(\widetilde{\mathcal{U}}^{(1)}, \widetilde{\mathcal{U}}^{(2)}, \dots, \widetilde{\mathcal{U}}^{(D)}\right).
$

Since $\widetilde{\mathcal{U}}_{:,i_k,:}^{(k)} = Q_{k-1}^\top\mathcal{U}_{:,i_k,:}^{(k)} Q_k$, $Q_k^\top Q_k = I$, and $Q_0,Q_D$ are identity matrices of appropriate sizes, we have:
\[
\begin{split}
    \mathcal{T}\left(\widetilde{\mathcal{U}}^{(1)}, \widetilde{\mathcal{U}}^{(2)}, \dots, \widetilde{\mathcal{U}}^{(D)}\right)_{\overline{i_1\dots i_D}, :} &= \left(Q_{0}^\top\mathcal{U}_{:,i_1,:}^{(1)} Q_1\right) \left(Q_{1}^\top\mathcal{U}_{:,i_2,:}^{(2)} Q_2\right)\dots \left(Q_{D-1}^\top\mathcal{U}_{:,i_D,:}^{(D)} Q_D\right) \\
    &= \mathcal{U}_{:,i_1,:}^{(1)}\mathcal{U}_{:,i_2,:}^{(2)}\dots \mathcal{U}_{:,i_D,:}^{(D)} = \mathcal{T}\left({\mathcal{U}}^{(1)}, \mathcal{{U}}^{(2)}, \dots, {\mathcal{U}}^{(D)}\right)_{\overline{i_1\dots i_D}, :}.
\end{split}
\]

Next, let us finally show that $\mathcal{M}(\widetilde{\mathcal{U}}^{(k)}) \in \mathrm{St}(R_{k-1}n_k, R_k)$.
Indeed, $\mathcal{M}(\widetilde{\mathcal{U}}^{(k)}) = (Q_{k\text{-}1}^\top \otimes I_{n_k}) \mathcal{M}(\mathcal{U}^{(k)}) Q_{k}$ since
\[
\begin{split}
    \left(\mathcal{M}(\widetilde{\mathcal{U}}^{(k)})\right)_{\overline{\alpha_{k-1} i_k}, \alpha_k}
    &=
    \sum_{\alpha_{k\text{-}1},\alpha_k=1}^{R_{k\text{-}1},R_k}
    (Q_{k\text{-}1})_{\alpha_{k\text{-}1},\beta_{k\text{-}1}}
    \mathcal{U}^{(k)}_{\alpha_{k\text{-}1},i_k,\alpha_k} (Q_{k})_{\alpha_{k},\beta_k}  \\
    & =\sum_{\alpha_{k\text{-}1},\alpha_k=1}^{R_{k\text{-}1},R_k} \sum_{j_k=1}^{n_k}
    (Q_{k\text{-}1})_{\alpha_{k\text{-}1},\beta_{k\text{-}1}}
    \delta_{i_k j_k}
    \mathcal{U}^{(k)}_{\alpha_{k\text{-}1},j_k,\alpha_k} (Q_{k})_{\alpha_{k},\beta_k}=
\end{split}
\]
\[
\begin{split}
    &= \sum_{\alpha_{k\text{-}1},\alpha_k=1}^{R_{k\text{-}1},R_k} \sum_{j_k=1}^{n_k}
    (Q_{k\text{-}1})_{\alpha_{k\text{-}1},\beta_{k\text{-}1}}
    \delta_{i_k j_k}
     \left(\mathcal{M}(\mathcal{U}^{(k)})\right)_{\overline{\alpha_{k-1} j_k}, \alpha_k}
     (Q_{k})_{\alpha_{k},\beta_k} \\
     &  
     =\left((Q_{k\text{-}1}^\top \otimes I_{n_k}) \mathcal{M}(\mathcal{U}^{(k)}) Q_{k}\right)_{\overline{\alpha_{k-1} i_k}, \alpha_k}.
\end{split}
\]
Hence,
\[
\begin{split}
    &\mathcal{M}(\widetilde{\mathcal{U}}^{(k)})^\top \mathcal{M}(\widetilde{\mathcal{U}}^{(k)}) 
    = 
    \left((Q_{k\text{-}1}^\top \otimes I_{n_k}) \mathcal{M}(\mathcal{U}^{(k)}) Q_{k}\right)^\top 
    \left((Q_{k\text{-}1}^\top \otimes I_{n_k}) \mathcal{M}(\mathcal{U}^{(k)}) Q_{k}\right) 
    \\
    &=Q_{k}^\top \mathcal{M}(\mathcal{U}^{(k)})^\top (Q_{k\text{-}1} \left(Q_{k\text{-}1}^\top)\otimes I_{n_k}\right) \mathcal{M}(\mathcal{U}^{(k)}) Q_{k}
    = Q_{k}^\top \mathcal{M}(\mathcal{U}^{(k)})^\top \mathcal{M}(\mathcal{U}^{(k)}) Q_{k} 
    = Q_{k}^\top Q_{k}  = I_{R_k},
\end{split}
\]
which completes the proof.

\section{STTP Degrees of Freedom}
\label{sec:sttp:dof}

We consider a tensor diagram from Fig.~\ref{fig:tt_network} made compatible with the TT decomposition introduced in Sec.~\ref{sec:prelim} (consisting only of TT-cores) by contracting the matrix $\Sigma$ into either left or right adjacent TT-core and recovering size-1 legs on the outer-most TT-cores. 
Such tensor diagram will have the following TT-rank and dimensions~\eqref{eq:ttdof}:
\begin{equation} 
\label{eq:doftt:sup}
\begin{split}
    &R=(1, R_1^{\dout}, \ldots, R_{D_{\dout}-1}^{\dout}, r
    , R_{D_{\din}-1}^{\din}, \ldots, R_1^{\din}, 1),
     \\
    &n=(n_1^{\dout}, \ldots, n_{D_{\dout}}^{\dout}, n_{D_{\din}}^{\din}, \ldots, n_1^{\din}).
\end{split}
\end{equation}

Recall that $R$ and $n$ are indexed in the ranges $[0,D_\dout + D_\din]$ and $[1,D_\dout + D_\din]$ respectively (Sec.~\ref{sec:prelim}) and that STTP is obtained by parameterizing its $\Sigma$ and TT-cores as follows (left to right in Fig.~\ref{fig:tt_network}, Sec.~\ref{sec:param_stiefel_tt}):

\begin{itemize}
    \item $\mathcal{M}(\mathcal{U}^{(k)}) \in \mathrm{St}_\mathsf{U}(R_{k-1}^{\dout} n_k^{\dout} \times R_k^{\dout}), 1 \leq k < D_\dout$, or equivalently $\mathrm{St}_\mathsf{U}(R_{k-1} n_k \times R_k), 1 \leq k < D_\dout$ (TT-cores of $U$ excluding the last one) in the notation~\eqref{eq:doftt:sup}, parameterized by $R_{k-1} n_k R_k - R_k^2$ parameters,
    \item $\mathcal{M}(\mathcal{U}^{(k)}) \in \mathrm{St}(R_{k-1}^{\dout} n_k^{\dout} \times R_k^{\dout}), k = D_\dout$, or equivalently $\mathrm{St}(R_{k-1} n_k \times R_k), k = D_\dout$ (the last TT-core of $U$) in the notation~\eqref{eq:doftt:sup}, parameterized by $R_{k-1} n_k R_k - R_k(R_k + 1)  \  / \  2$ parameters,
    \item $\Sigma = \mathrm{diag}(\sigma_1,\dots, \sigma_{r})$ is a matrix of singular values, parameterized by $r$ parameters,
    \item $\mathcal{M}(\mathcal{V}^{(k)}) \in \mathrm{St}(R_{k-1}^{\din} n_k^{\din} \times R_k^{\din}), k = D_\din$, or equivalently $\mathrm{St}(R_k n_k \times R_{k-1}), k = D_\dout + 1$ (the last TT-core of $V$) in the notation~\eqref{eq:doftt:sup}, parameterized by $R_k n_k R_{k-1} - R_{k-1} (R_{k-1} + 1) \  / \  2$ parameters,
    \item $\mathcal{M}(\mathcal{V}^{(k)}) \in \mathrm{St}_\mathsf{U}(R_{k-1}^{\din} n_k^{\din} \times R_k^{\din}), D_\din > k \geq 1$, or equivalently $\mathrm{St}_\mathsf{U}(R_k n_k \times R_{k-1}), D_\dout + 1 < k \leq D_\dout + D_\din$ (TT-cores of $V$ excluding the last one) in the notation~\eqref{eq:doftt:sup}, parameterized by $R_k n_k R_{k-1} - R_{k-1}^2$ parameters.
\end{itemize}

Summing up the degrees of freedom of STTP components listed above, and keeping in mind that $R_{D_\dout} \equiv r$,
\[
\begin{split}
  \mathrm{DOF}(W) 
  &= \left[ \sum_{k=1}^{D_\dout-1} R_{k-1} n_k R_k - R_k^2 \right] 
  + \left[ R_{k-1} n_k R_k - \frac{R_k (R_k + 1)}{2} \ \Big|\  k=D_\dout \right] 
  + r \ + \\
  & + \left[ R_k n_k R_{k-1} - \frac{R_{k-1} (R_{k-1} + 1)}{2} \ \Big|\  k=D_\dout + 1 \right] 
  + \left[ \sum_{k=D_\dout+2}^{D_\dout + D_\din} R_k n_k R_{k-1} - R_{k-1}^2 \right] = \\
  & =
  \left[ 
  r - \frac{r(r+1)}{2} - \frac{r(r+1)}{2}
  \right]
  + \sum_{k=1}^{D_\dout + D_\din} R_{k-1} n_k R_k
  - \sum_{\substack{k \in [1, D_\dout-1] \cup \\ [D_\dout + 1, D_\dout + D_\din - 1]}} R_k^2 \\
  & = 
  \sum_{k=1}^{D_\dout + D_\din} R_{k-1} n_k R_k
  - \sum_{k=1}^{D_\dout + D_\din-1} R_k^2,
\end{split}
\]
we arrive at the same dimensionality of the fixed TT-rank tensor manifold as \cite{holtz2012manifolds}.

{
\bibliographystyle{biochem}
\bibliography{all}
}

\end{document}